%% file: main.tex
\documentclass[11pt]{article}

\usepackage{amssymb,amsmath,amsthm,amsfonts}

\usepackage[T1]{fontenc}
\usepackage{times}

\usepackage[margin=1in]{geometry}
\usepackage[ruled]{algorithm2e} 
    
    \SetAlFnt{\small}
    \SetAlCapFnt{\small}
    \SetAlCapNameFnt{\small}
    \SetAlCapHSkip{0pt} 
    \IncMargin{-\parindent}
\newenvironment{algorithm2e}[1][t]{%
    \begin{algorithm}[#1]%
}{%
    \end{algorithm}%
}

\usepackage{enumitem}
\setlist{nosep,topsep=0pt,leftmargin=*}

\usepackage{xspace,nicefrac,graphicx,tcolorbox,xifthen,thm-restate,tikz,hyperref,booktabs}

\usepackage[bb=dsserif]{mathalpha}
\hypersetup{
    colorlinks,
    allcolors=myblue
}

\usepackage{xcolor}
    \definecolor{myred}{HTML}{ea4335}
    \definecolor{mygreen}{HTML}{41a756}
    \definecolor{myblue}{HTML}{4285f4}

\usepackage[capitalize]{cleveref}

\usepackage[sortcites,sorting=nyt,style=alphabetic]{biblatex}
\newtheorem{theorem}{Theorem}[section]

\newtheorem{lemma}[theorem]{Lemma}

\newtheorem{remark}[theorem]{Remark}

\theoremstyle{definition}

\newtheorem{definition}{Definition}[section]

\let\a\alpha
\let\g\gamma
\let\e\varepsilon
\let\d\delta
\let\l\lambda

\newcommand{\term}[1]{\ensuremath{\mathtt{#1}}\xspace}

\newcommand{\reg}{\term{Reg}}
\newcommand{\Amax}{\term{Alg_{max}}}
\newcommand{\Amin}{\term{Alg_{min}}}
\newcommand{\opt}{\term{OPT}}
\newcommand{\optfd}{\term{OPT_{FD}}}

\newcommand{\rew}{\term{REW}}
\newcommand{\mab}{\term{MAB}}
\newcommand{\bwk}{\term{BwK}}
\newcommand{\sR}{\sigma_{\mathtt{r}}}
\newcommand{\sC}{\sigma_{\mathtt{c}}}
\newcommand{\hist}{\mathcal{H}}

\newcommand{\R}{\mathbb{R}}

\newcommand{\calL}{\mathcal{L}}
\newcommand{\TA}{T_\mathtt{A}}
\newcommand{\Tres}{T_\mathtt{res}}

\newcommand{\Rmin}{\reg_{\min}}
\newcommand{\Rmax}{\reg_{\max}}

\newcommand{\Ex}[2][]{\mathop{\mathbb{E}}_{#1}\left[#2\right]}
\renewcommand{\Pr}[2][]{\mathbb{P}_{#1}\left[#2\right]}
\newcommand{\One}[1]{\mathbb{1}\left[#1\right]}

\newcommand{\Line}[4]{%
    #1 &\;#2\;#3\ifthenelse{\isempty{#4}}{}{&&\qquad\left(#4\right)}
}

\newcommand{\citet}[1]{\textcite{#1}}

\title{Approximately Stationary Bandits with Knapsacks}

\author{
    Giannis Fikioris
    \thanks{Supported by the Department of Defense (DoD) through the National Defense Science \& Engineering Graduate (NDSEG) Fellowship, the Onassis Foundation -- Scholarship ID: F ZS 068-1/2022-2023, and AFOSR grant FA9550-23-1-0068.}\\
    Cornell University\\
    \texttt{gfikioris@cs.cornell.edu}
    \and
    \'Eva Tardos
    \thanks{Supported in part by AFOSR grant FA9550-19-1-0183 and FA9550-23-1-0068.}\\
    Cornell University\\
    \texttt{eva.tardos@cornell.edu}
}

\date{\vspace{-25pt}}

\begin{document}

\maketitle

\begin{abstract}
    \input{body/0.abstract}
\end{abstract}

\input{body/1.intro}
\input{body/2.related}
\input{body/3.preliminaries}
\input{body/4.stationary}
\input{body/5.algo_analysis}
\input{body/6.simple_guarantees}
\input{body/7.impossibility}
\input{body/8.complex_guarantees}

\printbibliography

\appendix
\input{appendix/6.simple}
\input{appendix/7.impossibility}
\input{appendix/8.complex}

\end{document}

%% file: body/0.abstract.tex
\textit{Bandits with Knapsacks} (\bwk), the generalization of the \textit{Multi-Armed Bandits} problem under global budget constraints, has received a lot of attention in recent years. It has numerous applications, including dynamic pricing, repeated auctions, ad allocation, network scheduling, etc. Previous work has focused on one of the two extremes: \textit{Stochastic \bwk} where the rewards and consumptions of the resources of each round are sampled from an i.i.d. distribution, and \textit{Adversarial \bwk} where these parameters are picked by an adversary. Achievable guarantees in the two cases exhibit a massive gap: No-regret learning is achievable in the stochastic case, but in the adversarial case only competitive ratio style guarantees are achievable, where the competitive ratio depends either on the budget or on both the time and the number of resources. What makes this gap so vast is that in Adversarial \bwk the guarantees get worse in the typical case when the budget is more binding. While ``best-of-both-worlds'' type algorithms are known (single algorithms that provide the best achievable guarantee in each extreme case), their bounds degrade to the adversarial case as soon as the environment is not fully stochastic.

Our work aims to bridge this gap, offering guarantees for a workload that is not exactly stochastic but is also not worst-case. We define a condition, \textit{Approximately Stationary \bwk}, that parameterizes how close to stochastic or adversarial an instance is. Based on these parameters, we explore what is the best competitive ratio attainable in \bwk. We explore two algorithms that are oblivious to the values of the parameters but guarantee competitive ratios that smoothly transition between the best possible guarantees in the two extreme cases, depending on the values of the parameters. Our guarantees offer great improvement over the adversarial guarantee, especially when the available budget is small. We also prove bounds on the achievable guarantee, showing that our results are approximately tight when the budget is small.

%% file: body/1.intro.tex
\section{Introduction}
\label{sec:intro}

\textit{Bandits with Knapsacks} (\bwk) was first introduced in \cite{DBLP:conf/focs/BadanidiyuruKS13} and models a natural extension of the \textit{Multi-Armed Bandit} (\mab) problem. In \mab a player repeatedly chooses one of many actions, each providing an unknown reward. To maximize her total reward, a player needs to balance exploration and exploitation while picking her actions. In the budgeted version of the problem (\bwk), the player has the same objective but also needs to be mindful of different resources: each action consumes some amount of each resource; if any resource is depleted the player cannot get any more rewards. \bwk was initially formulated inspired by numerous practical problems where a player wants to maximize her reward with constraints: participating in repeated auctions, dynamic pricing, ad allocation, network routing/scheduling, etc.

Previous work on \bwk has focused on two extreme cases of the problem. First, in \textit{Stochastic \bwk} the environment (rewards and consumptions) in each round is sampled from a distribution identical and independent of other rounds. Second, in \textit{Adversarial \bwk} the environment is picked each round by an adversary. Unlike \mab, in \bwk there is a clear dichotomy in guarantees between the two cases. No-regret learning, i.e., additive error sublinear in the total number of rounds, is achievable in the stochastic case, but not in the adversarial case. Instead, work in Adversarial \bwk focuses on bounding the achievable \textit{competitive ratio}, i.e., the multiplicative error on the achievable reward. A line of work that tries to connect the adversarial and stochastic cases is ``best-of-both-worlds'' type results, where an algorithm achieves guarantees in both settings, without knowing if the environment is adversarial or stochastic, e.g., see \cite{DBLP:conf/icml/CastiglioniCK22}. However, the guarantee offered degrades to the adversarial guarantee as soon as the setting is not fully stochastic.

In this work, we aim to bridge this vast gap between Stochastic and Adversarial \bwk and offer performance guarantees that smoothly degrade depending on the deviation from stochasticity, extending the ``best-of-both-words'' style guarantees to cases between the two extremes.
We call our framework \textit{Approximately Stationary \bwk}, which offers a natural interpolation between Stochastic and Adversarial \bwk. Adversarial \bwk is much harder because of the potential for \textit{huge heterogeneity of environments between rounds}. In the Approximately Stationary \bwk problem, we limit this heterogeneity by assuming that the change in \textit{expected} rewards and consumptions of any arm is limited. We assume that the player is not aware of the parameters limiting the change while running the algorithm. A natural constraint is that if $x_t$ is the expected reward of some action in round $t$, it must hold that $\min x_t \ge \sigma \max x_t$, with $\sigma$ limiting the variability of the expectation.
In Stochastic \bwk it must hold that $\sigma = 1$ (actually, distributions are identical across rounds, not just expectations), and in Adversarial \bwk it can be that $\sigma = 0$.

There are multiple settings where the value of $\sigma$ is neither of the two extremes. Consider repeated auctions where every round a budget-limited player bids to win a certain item. If the player's value for the item and its price are independently and identically distributed across rounds, then the setting is completely stochastic, and $\sigma = 1$. However, this is rarely the case in practice. The distribution of values might change across rounds (e.g., seasonal differences) or the price might be controlled by other players' bids who change their strategy or by a central entity that lowers or raises the price. This means that $\sigma < 1$, but the values and prices are not adversarial or arbitrary, i.e., the variability of the price is limited. 
Our goal is to have an algorithm that will have the best guarantees given the value of $\sigma$, without knowing its actual value. 
In this paper, we show that it is possible to achieve performance that degrades gradually depending on the value $\sigma$. For the most interesting range of the player's budget, we obtain close to optimal performance for all values of $\sigma$ without assuming that the player is aware of its value.

\paragraph{Overview of our results.}
We introduce our model, \textit{Approximately Stochastic \bwk}, in Section~\ref{sec:stat}. Our model interpolates between Stochastic and Adversarial \bwk by having two parameters that limit how much the expected value of the rewards and consumptions of any action can change across rounds.
More specifically, the parameter $\sR \in [0,1]$ limits how much the reward of any action can fluctuate across rounds: if $r_t(a)$ is the \textit{expected reward} of action $a$ in round $t$, we require that $\min_t r_t(a) \ge \sR \max_t r_t(a)$ for all actions $a$.
Note that we apply this definition to the expected reward of an arm since even in Stochastic \bwk the observed rewards of a single arm can range from $0$ to $1$. 
Similarly, the parameter $\sC \in [0,1]$ limits the consumptions of any action across rounds: if $c_{t, i}(a)$ is the \textit{expected consumption} of resource $i$ by action $a$ in round $t$, we require that $\min_t c_{t, i}(a) \ge \sC \max_t c_{t, i}(a)$ for all actions $a$ and resources $i$.
A sequence of rewards and consumptions that satisfy the above constraints is called \textit{$(\sR, \sC)$-stationary}.
Aside from this restriction, we make no assumptions, so we think that the adversary can be adaptive, i.e., pick the rewards and consumptions of a round based on the past rounds.
We note that in Stochastic \bwk the adversary is $(1,1)$-stationary and in Adversarial \bwk the adversary is $(0, 0)$-stationary.
Our framework naturally generalizes ``best-of-both-worlds'' approaches, offering guarantees not only in fully stochastic and adversarial environments but also environments between the two extremes.
We note that using two parameters instead of one that constrains both rewards and consumptions is valuable: a small $\sR$ has a much different effect on the achievable guarantees than a small $\sC$.

As is standard we assume without loss of generality that the rewards and consumptions are non-negative and bounded by $1$ and every resource has budget $B$.
In adversarial \bwk the best competitive ratio depends on the player's average (or per round) budget. We will use $\rho = B/T$ to denote the player's per-round budget.
Given that the consumptions are bounded by $1$ each round, $\rho=1$ means that the player is not budget limited; in more typical cases players have budgets only for a small fraction of the items available.
Our goal is to design algorithms that guarantee a fraction of the optimal solution when the player's per-round budget is $\rho$ and the adversary is $(\sR, \sC)$-stationary. We denote this fraction with $\a_\rho(\sR, \sC)$. There are algorithms that have $\a_\rho(1, 1) = 1$ and $\a_\rho(0, 0) = \rho$, and these are best possible when $\rho$ is a constant independent of the time horizon, but nothing is known for intermediate values. This effectively means that previous work can only guarantee $\a_\rho(\sR, \sC) = \rho$ when $\sR < 1$ or $\sC < 1$. This is an enormous and unnatural gap, especially in the typical case when $\rho$ is small. We study algorithms that are oblivious to the values of $\sR$ and $\sC$ and achieve a fraction $\a_\rho(\sR, \sC)$ of the optimal solution that is continuous and increasing in both arguments and satisfies $\a_\rho(1, 1) = 1$ and $\a_\rho(0, 0) = \rho$.

All of our guarantees are against an \textit{adaptive adversary}, an adversary that is restricted to be $(\sR, \sC)$-stationary, but beyond this restriction, is allowed to pick the distributions of rewards and consumptions of each round based on outcomes in previous rounds.
This is in contrast to an \textit{oblivious adversary}, who picks all the rewards and consumptions upfront.
Allowing an adaptive adversary is important: it makes our guarantees apply when the algorithm is used in a multi-agent game setting, e.g. in repeated auctions, where prices depend on other agents' bids, who are all adaptive to the history of play.
All the guarantees we present in this paper are against an adaptive adversary, which, to the best of our knowledge, are the first such guarantees for Adversarial \bwk.

In Section~\ref{sec:guar} we present our first guarantee. We show that we can achieve a guarantee of $\a_\rho(\sR, \sC) \ge \rho + \sR(\sC - \rho)^+$ (Theorem~\ref{thm:guar:guar}). The most interesting range of parameters is when $\rho$ is small (and therefore the gap between Stochastic and Adversarial \bwk is biggest) and $\sR\sC$ is much bigger than $\rho$. In this case, our guarantee becomes approximately $\sR \sC$. This is in stark contrast to the guarantee that previous work would suggest, $\rho$. An alternate, naive approach would be to use an algorithm assuming a fully stochastic setting, which would yield a guarantee of $\approx \sR^2 \sC^2$ (see Section~\ref{sec:guar}), two orders of magnitude smaller. Our results show that even if the player has a small budget, as long as the environment is not completely adversarial, good guarantees are achievable. We note that a small budget is indeed the most common case: Typical budgets are far from sufficient for all items, so $\rho$ is small, and the environment is often less variable and independent of the player's budget. For example, in repeated auctions, we expect that expected item prices fluctuate across rounds but this fluctuation to not depend on the player's small budget.

In Section~\ref{sec:imposs} we provide a bound on the achievable guarantees any algorithm can get. The upper bound of Theorem~\ref{thm:imposs} on $\a_\rho(\sR, \sC)$ is using only one resource and an oblivious adversary.
When $\rho$ is much smaller than $\sR \sC^2$, Theorem~\ref{thm:imposs} shows that $\a_\rho(\sR, \sC) \lessapprox \sR \sC$, making the result of Theorem~\ref{thm:guar:guar} almost tight in that case. 
Theorem~\ref{thm:imposs} also shows that when $\sR = O(\rho)$ then $\a_\rho(\sR, \sC) \le O(\rho)$, making Theorem~\ref{thm:guar:guar} (and any algorithm with the guarantee $\a_\rho(\sR, \sC) \ge \rho$) tight up to constant factors.
Another interesting corollary of our theorem is that $\a_\rho(0, 0) \le \rho$, i.e., that a $\rho$ fraction of the optimal reward is the best possible in Adversarial \bwk\footnote{This bound was previously proven by \cite{DBLP:conf/sigecom/BalseiroG17} when the benchmark used for the optimal solution is the best sequence of actions. Our bound uses the best distribution of actions which is the standard benchmark in \bwk since the previous one is attainable only in special cases.}.
The bound of Theorem~\ref{thm:imposs} relies on that any algorithm needs to conserve its budget because of the uncertainty of the future. Without knowing the values of $\sR$ and $\sC$ it is not possible to utilize the reward of the optimal action in the initial rounds, in fear of the adversary being $(0, 0)$-stationary.

Our first guarantee in Theorem~\ref{thm:guar:guar} is close to optimal when $\sC$ is much bigger than $\rho$. When $\sC$ is similar to $\rho$ the guarantee is very small: if $\sC \le \rho$ the theorem only provides the adversarial guarantee of $\a_\rho(\sR, \sC) \ge \rho$.
In Section~\ref{sec:complex} we offer better guarantees for the case when $\sC$ and $\rho$ are small, but $\sR$ is larger.
Our improved guarantee requires an additional condition to approximate stationarity: the total per-round change of the consumption of any action is sublinear in the total number of rounds. Under this assumption and $(\sR, \sC)$-stationarity, we provide improved bounds in Theorem~\ref{thm:complex:guar}.
The improvement is most impressive when there is only one resource (e.g., money), $\sC$ is close to $\rho$, and $\sR$ is much larger, in which case the guarantee is close to the best possible (see Theorem~\ref{thm:imposs}).

Theorem~\ref{thm:guar:guar} is based on a simplified version of the algorithm in \cite{DBLP:conf/icml/CastiglioniCK22} (see Algorithm \ref{algo:1} in Section \ref{sec:algo}), while to achieve the guarantee of Theorem~\ref{thm:complex:guar} we need to break the time horizon into smaller batches and restart the algorithm periodically (see Algorithm \ref{algo:2} in Section \ref{sec:complex}).
The main technical lemmas leading to the guarantees in both cases follow an interesting and novel approach. They prove that the rewards of our algorithms have no-regret against the total reward of any action \textit{across all rounds} but scaled down to observe the average consumption bound in each iteration.
This property is not useful in Stochastic \bwk, where one knows that the optimal arm has low consumption, or in Adversarial \bwk, where the rewards of the optimal arm can become $0$ after a certain round.
In Approximately Stochastic \bwk, where the reward of the optimal arm cannot become $0$ (unless $\sR = 0$) but its consumption can increase, guaranteeing a fraction of the reward from the entire period is a very useful property.
The auxiliary lemma we use to prove the improved guarantee of Theorem~\ref{thm:complex:guar} scales down the reward of each round based on the consumption of the optimal arm \textit{in that round}. This makes its guarantee stronger when $\sC$ is small compared to the lemma used for Theorem~\ref{thm:guar:guar} where the reward needs to be scaled down using the maximum consumption across rounds.

%% file: body/2.related.tex
\paragraph{Related work.}
There is a vast amount of literature on online learning and regret minimization; we refer the reader to textbooks like \cite{DBLP:journals/ftopt/Hazan16} and \cite{DBLP:journals/sigecom/Slivkins20} for background. The most commonly used algorithms for \mab are Hedge for full information feedback (\cite{DBLP:journals/jcss/FreundS97}) and EXP3 for bandit feedback (\cite{DBLP:journals/siamcomp/AuerCFS02}). A ``best-of-both-worlds'' type result providing guarantees for both stochastic and adversarial \mab was first proven in \cite{DBLP:journals/jmlr/BubeckS12}. 

The \bwk framework was first introduced in \cite{DBLP:conf/focs/BadanidiyuruKS13} in the stochastic setting. Following this work, various extensions have been studied including concave rewards and convex consumptions (\cite{DBLP:conf/sigecom/AgrawalD14}), combinatorial semi-bandits with knapsacks (\cite{DBLP:conf/aistats/SankararamanS18}), contextual bandits (\cite{DBLP:conf/colt/BadanidiyuruLS14}), and negative consumptions that replenish the player's budget (\cite{DBLP:conf/nips/KumarK22}).

The first guarantees for Adversarial \bwk were by \cite{DBLP:conf/focs/ImmorlicaSSS19} and then by \cite{DBLP:conf/colt/Kesselheim020}. The first work guarantees a $1/O(d \log T)$ fraction of the optimal solution and the second a $1/O(\log d \log T)$ fraction, where $d$ is the number of resources. The second guarantee is tight when the average budget is $\rho = O(T^{-\e})$ for any constant $\e > 0$.
\cite{DBLP:conf/sigecom/BalseiroG17} study repeated second-price auctions with budgets, a special case of \bwk; they assume $\rho = \Omega(1)$, and get a ``best-of-both-worlds'' guarantee: a constant $\rho$ fraction of the optimal solution in adversarial settings and no-regret in stochastic ones. They also prove that even in this restricted second-price setting, their adversarial result is tight when $\rho$ is a constant and the comparator is the best sequence of actions, an extremely strong benchmark that an algorithm can be competitive against only in special cases of \bwk, like second-price auctions.
\cite{DBLP:conf/icml/CastiglioniCK22} generalize the previous ``best-of-both-worlds'' guarantee for general \bwk.
\cite{DBLP:conf/ijcai/RangiFT19} study a variant of \bwk where there is no time horizon and only one resource whose consumption is strictly positive; the game stops when that resource is depleted. Their variant is much easier than general \bwk, evident by their guarantees of $\textrm{poly}\log T$ and $O(\sqrt T)$ regret bounds for the stochastic and adversarial cases, respectively; in general \bwk these types of guarantees are not achievable.

There have been other models that interpolate between Stochastic and Adversarial \mab.
\cite{DBLP:conf/stoc/LykourisML18} study Stochastic \mab with Corruptions where the environment is stationary except for $C$ rounds that are corrupted by an adversary; their no-regret guarantee interpolates between the stochastic and adversarial guarantees depending on how large $C$ is.
In Restless Bandits, e.g., studied by \cite{DBLP:journals/tit/TekinL12} and \cite{DBLP:conf/nips/WangHL20}, the environment each round is generated by a Markov Chain whose state changes each round. Algorithms have no-regret when the Markov chain has size much smaller than $T$, making the environments of two rounds approximately independent if they are far enough apart.
Both of these models interpolate between the adversarial and stochastic case but in a fundamentally different way than ours: Before round $1$ both have an ``expected environment'' (the uncorrupted one in Stochastic \mab with Corruptions and the one generated by the stationary distribution of the Markov chain in Restless bandits) which the environments of most rounds are ``close'' to in expectation. Instead, our $(\sR, \sC)$-adversary does not have to conform to this restriction and can vary environments significantly at each time step.

Previous work has considered \mab with close to stationarity constraints similar to ours.
\cite{DBLP:conf/nips/GurZB14} impose a constraint on the rewards of the actions similar to the one we consider in Section \ref{sec:complex}: they bound the total per-round change of the reward of any action by a parameter $V$.
They prove regret bounds of order $\tilde \Theta(V^{1/3} T^{2/3})$ against an oblivious adversary.
\cite{DBLP:journals/ior/BesbesGZ15} generalize the previous results for general action spaces and convex reward functions.
In contrast, having only this constraint for both the rewards and consumptions of the actions in \bwk, even with $V = 1$, does not improve over Adversarial \bwk: Theorem~\ref{thm:imposs} satisfies the above constraint with $\sR = \sC = 0$ which shows that only a $\rho$ fraction of the optimal solution is achievable.

No-regret guarantees have been studied for \bwk under stationarity conditions similar the one above. \cite{DBLP:conf/nips/LiuJL22} design no-regret algorithms when the reward and consumptions of every action change slowly over time and remain close to their average value (with known bounds for these constraints). 
Concurrently and independently with our work, \cite{slivkins2023contextual} study stochastic \textit{contextual bandits with linear constraints} (a generalization of contextual \bwk) and design no-regret algorithms for this setting. They also consider environments that are mostly stationary with a small number of changes in the distribution of rewards and consumptions.
In this environment, if the number of changes is not too big, they achieve no-regret against a benchmark similar to the one we use in Lemma~\ref{lem:compex} but make no guarantees against the standard benchmark we use in our theorems.
The results of both \cite{DBLP:conf/nips/LiuJL22} and \cite{slivkins2023contextual} are incomparable to the ones we make for Approximately Stationary \bwk: even if $\sR = \sC = 1 - \e$ for some small constant $\e$, the variability in the environment can be too high to yield meaningful regret guarantees.

\textit{Online allocation problems} are similar to \bwk except that the player gets to observe the rewards and consumptions of a round \textit{before} she picks an action for that round and so the only uncertainty comes from future rewards and consumptions. \citet{DBLP:journals/jacm/DevanurJSW19} prove no-regret bounds for this problem in the stochastic setting. They also study the adversarial case where their algorithm guarantees no-regret against a specific benchmark as long as their algorithm knows the value of this benchmark. More specifically, they compare against $\min_t \opt_t$, where $\opt_t$ is the maximum achievable reward if the rewards and consumptions of all rounds were distributed similarly to those of round $t$ (which can be adversarially picked).
\cite{DBLP:conf/icml/BalseiroLM20} study the same problem and provide guarantees similar to \cite{DBLP:conf/icml/CastiglioniCK22} in the stochastic and adversarial cases. Additionally, they study models that interpolate the two extremes, similar to Stochastic \mab with Corruptions and Restless Bandits. Their results, similar to \cite{DBLP:journals/jacm/DevanurJSW19}, focus on getting no-regret guarantees in those models.

%% file: body/3.preliminaries.tex
\section{Bandits with Knapsacks}

In this section, we formally define the Bandits with Knapsack (\bwk) framework.
Our notation and definitions are similar to \cite[Chapter 10]{DBLP:journals/sigecom/Slivkins20}. We introduce some additional notation to help distinguish between the realized reward/consumptions of an action and their expectation. 

There are $T$ rounds, $d$ \textit{resources} (denoted with $[d]$), and a \textit{budget} per resource which w.l.o.g. we assume is the same for every resource $i\in [d]$, $B$. We denote $\rho = B/T$. The player has a set of $K$ \textit{actions}, $[K]$. In every round $t\in [T]$ the adversary chooses $(d+1)K$ random (and potentially dependent on previous rounds and of each other) variables: $R_t, C_{t,1}, \ldots, C_{t,d} : [K] \to [0,1]$.
For an action $a\in [K]$, $R_t(a)$ is the \textit{reward} the player receives on round $t$ if they play action $a$, and for a resource $i \in [d]$, $C_{t,i}(a)$ is the \textit{consumption} of that resource in round $t$ by $a$.
Because consumptions are upper bounded by $1$ we can assume that $\rho \le 1$.
As is standard, we also assume that there is an action, called the \textit{null action}, with $0$ reward and $0$ consumption of every resource.

Every round $t$ the player chooses a (potentially randomized) action $A_t \in [K]$ without any knowledge of the reward or consumptions of that round or their distribution. The game ends either after round $T$ or in the round that any resource is depleted. We define $\TA$ to be the last round the player receives a reward:
$\TA = \max\left\{
      t \in [T] : \forall i\in [d], \sum_{\tau=1}^t C_{\tau,i}(A_\tau) \le B
   \right\}
$. 

We denote with $\hist_t$ the \textit{history} up to round $t$. This includes the realization of the actions of the player up to round $t$, $A_1, \ldots, A_t$, as well as the realization of the rewards and consumptions up to round $t$, $R_1, \ldots, R_t$ and $\{C_{1, i}\}_i, \ldots, \{C_{t, i}\}_i$. We generally assume that the player has bandit knowledge, i.e., only knows the realization of rewards and consumptions of the actions she took in previous rounds. As we note later, our results can be improved if the player has full information knowledge.

If for every round $t$ the distributions of the functions $R_t, C_{t,1}, \ldots, C_{t,d}$ are independent of $\hist_{t-1}$ (which allows dependence on the algorithm used by the player but not the realization of its actions) then the adversary is called \textit{oblivious}; otherwise, the adversary is called \textit{adaptive}. Additionally, if the aforementioned functions are not time dependant, the adversary is called \textit{stochastic}.

For any action $a\in [K]$, we denote with $r_t(a)$ the expected reward of action $a$ in round $t$ \textit{conditioned on the history of the previous rounds}: $r_t(a) = \Ex{R_t(a) | \hist_{t-1}}$. Similarly, we define the conditional expected cost of action $a$ and resource $i$: $c_{t, i}(a) = \Ex{C_{t, i}(a) | \hist_{t-1}}$.

%% file: body/4.stationary.tex
\section{Approximately Stationary Bandits with Knapsacks}
\label{sec:stat}

In this section, we present our model, \textit{Approximately Stationary \bwk}. Our model interpolates between Stationary and Adversarial \bwk, providing guarantees that smoothly improve as expectations change less across time or equivalently as the setting is less adversarial. This generalizes ``best-of-both-worlds'' results by providing guarantees for the whole spectrum, not just the extremes.

As we mentioned in the introduction, Adversarial \bwk is hard because the rewards and consumptions of an arm can oscillate between extreme values. The issue with Adversarial \bwk becomes apparent by looking at the impossibility results of \cite{DBLP:conf/sigecom/BalseiroG17} and \cite{DBLP:conf/focs/ImmorlicaSSS19}. Both have similar structure: there is only one action with positive reward every round. However, even if the player knows which action that is, she cannot fully utilize it, in case some other action has a much larger reward in a later round.
Our model limits these extreme-case examples by constraining the expectation of the rewards and consumptions the adversary can pick. We focus on expected rewards and consumptions since even in Stochastic \bwk the range of the realized rewards can be the entire interval $[0, 1]$. Our definition uses two parameters. The first, $\sR$, bounds the multiplicative difference in the maximum and minimum expected reward of any action. The second, $\sC$, similarly constrains the expected consumption of any resource by any action.

\begin{definition}\label{def:stat_bwk}
    An adversary in \bwk is called \textit{$(\sR, \sC)$-stationary} if for any action $a$, resource $i$, and history $\hist_T$ it holds that $\min_t r_t(a) \ge \sR \max_t r_t(a)$ and $\min_t c_{t, i}(a) \ge \sC \max_t c_{t, i}(a)$.
\end{definition}

\begin{remark}
    First, note that our definition bounds the relative variation of a sequence and not the absolute one, e.g., $\max_t r_t(a) - \min_t r_t(a) \le \e$. Bounding the relative size rather than the absolute difference makes sense, making the result invariant on the scale of the rewards.
    Second, we notice that if $\sR = \sC = 0$ then the setting is completely adversarial. If the setting is Stochastic \bwk, we get $\sR = \sC = 1$.
    Third, since the reward constraint is applied to the expected rewards of an action \textit{given the history of the previous rounds}, our adversary can be adaptive.
\end{remark}

%% file: body/5.algo_analysis.tex
\section{Benchmarks and Algorithm}
\label{sec:algo}

In this section, we present the benchmark that we use to compare the quality of our algorithm, as well as the algorithm that we use that provides our first guarantee.

\paragraph{Benchmark.}
The benchmark we use is the standard \textit{best-fixed distribution of actions in hindsight}. Its reward $\optfd$ is equal to the reward of the best distribution of actions $A^* \in \Delta([K])$, up to the round when it runs out of budget. For simplicity of presentation, we define $\optfd$ using the expected rewards and consumptions $r,c$:
\begin{equation}\label{eq:optfd}
\begin{split}
    \optfd
    =
    \max_{\substack{T^*\in [T] \\ A^* \in \Delta([K])}} &\quad
    \sum_{t=1}^{T^*} \Ex[a\sim A^*]{r_t(a)}
    \\
    \textrm{such that} &\quad
    \sum_{t=1}^{T^*} \Ex[a\sim A^*]{c_{t,i}(a)} \le B, \quad \forall i\in [d]
\end{split}
\end{equation}
We note that in $\Ex[a\sim A^*]{\cdot}$ the expectation is taken only over the action $a\sim A^*$ and not any choices the player or adversary make, i.e., $\Ex[a\sim A^*]{r_t(a)} = \sum_a \Pr{A^* = a} r_t(a)$, where $r_t(a)$ is the expected reward of action $a$ for the actual history of the play, and not for the history of playing action distribution $A^*$ each round. This means that $\optfd$ depends on the realization of the random choices of the game, i.e., the player's and adversary's actions. This is similar to benchmarks in \mab with an adaptive adversary, where the optimal reward used as a comparator for no-regret depends on the actions the player takes.
We denote with $(T^*, A^*)$ the solution to the above optimization problem.

Optimization problem \eqref{eq:optfd} is simplified when the expectations of the rewards and consumptions are the same every round $r_t(\cdot) = r(\cdot)$ and $c_{t, i}(\cdot) = c_i(\cdot)$ for all $t, i$. In this case, \eqref{eq:optfd} becomes
\begin{equation}\label{eq:optfd_stoch}
    \max_{A^* \in \Delta([K])} \quad
    T \Ex[a\sim A^*]{r(a)}
    \qquad
    \textrm{such that} \quad
    T \Ex[a\sim A^*]{c_i(a)} \le B, \quad \forall i\in [d]
\end{equation}
where we can drop the dependence on $T^*$ because of the null action: For every feasible solution $(\hat T, \hat A)$ there is a feasible solution $(T, \hat A')$ with the same reward. More specifically, $\hat A'$ is the same distribution as $\hat A$ with probability $\hat T/T$ and the null action otherwise. 

We use $\rew$ to denote the total reward of the player: $\rew = \sum_{t=1}^{T_A} R_t(A_t)$. We focus on high-probability bounds.
The player has competitive ratio $\g \ge 1$ and regret $\reg$ against $\optfd$ with probability $1-\d$ if $\Pr{ \rew \ge \frac{\optfd - \reg}{\g} } \ge 1-\d$.

As mentioned in the related work, in Adversarial \bwk the player can guarantee competitive ratio $\g = \min\{1/\rho, O(\log T)\}$ and sublinear regret. Without any additional assumptions, this result is tight. By assuming that the adversary is $(\sR, \sC)$-stationary, we prove greatly improved guarantees for the competitive ratio $\g$. Our guarantees provide a smooth interpolation between Adversarial and Stochastic \bwk.

\paragraph{Algorithm based on Lagrangian maximization/minimization.}
Next, we present a simplified version of the algorithm of \cite{DBLP:conf/icml/CastiglioniCK22} that achieves a ``best-of-both-worlds'' guarantee: competitive ratio of $1$ in stochastic environments and $1/\rho$ in adversarial ones. Our first guarantee against a $(\sR, \sC)$-adversary in Theorem~\ref{thm:guar:guar} is based on this algorithm and provides a competitive ratio that ranges between $1$ and $1/\rho$ depending on the values of $\sR$ and $\sC$.

The algorithm is inspired by the Lagrangian of \eqref{eq:optfd_stoch},
$\calL(a, \vec \l) = r(a) + \sum_{i\in [d]}\l_i (\rho - c_i(a)) $
where $\l \in \R_{\ge 0}^d$. The importance of this function can be seen by the fact that in the stochastic case
\begin{align}\label{eq:algo:minmax}
    \frac{\optfd}{T}
    =
    \max_{A \in \Delta([K])} \min_{\vec\l \in \R_{\ge 0}^d} \Ex[a\sim A]{\calL(a, \vec \l)}
    =
    \min_{\vec\l \in \R_{\ge 0}^d} \max_{A \in \Delta([K])} \Ex[a\sim A]{\calL(a, \vec \l)}
\end{align}
as shown by by \cite{DBLP:conf/focs/ImmorlicaSSS19}.
\cite{DBLP:conf/icml/CastiglioniCK22} improve \eqref{eq:algo:minmax} by restricting the domain of $\vec\l$: \eqref{eq:algo:minmax} also holds if the minimum is over $\vec\l \in \mathcal D$ where $\mathcal D = \{\vec\l \in \R_{\ge 0}^d : \sum_i \l_i \le 1/\rho\}$. 

Even though the original inspiration comes from Stochastic \bwk, previous work designed Adversarial and Stochastic \bwk algorithms that aim to find a saddle point of a Lagrangian.
However, the player does not know the expected values of the rewards and consumptions, so their realized values are used instead. Additionally, since the Lagrangian is linear in $\vec \l$, similarly to \cite{DBLP:conf/focs/ImmorlicaSSS19}, we replace the domain $\mathcal D$ with its extreme points: instead of $\vec \l$ the second argument becomes $i \in [d]\cup\{0\}$ where $i=0$ corresponds to the zero vector and $i > 0$ corresponds to the all-zero vector with $1/\rho$ in its $i$-th position. Handling this function is easier since it is defined over a discrete set. Putting all of these together, we define for every $t\in [T]$, $a\in [K]$, and $i\in [d]\cup\{0\}$:
\begin{align*}
    \calL_t(a, i)
    =
    R_t(a) + \frac{1}{\rho}\One{i \ne 0}(\rho - C_{t,i}(a))
    .
\end{align*}

The algorithms of \cite{DBLP:conf/focs/ImmorlicaSSS19} and \cite{DBLP:conf/icml/CastiglioniCK22} focus on finding a saddle point of functions similar to the above\footnote{\cite{DBLP:conf/icml/CastiglioniCK22} use the initial formulation with $\vec \l\in\mathcal D$. In \cite{DBLP:conf/focs/ImmorlicaSSS19} the second argument of the function considers only the $d$ non-zero extreme points, i.e., the domain of $i$ is $[d]$ instead of $[d]\cup\{0\}$; even if every consumption is less than $\rho$ (in which case the algorithm is not budget constrained) the choice of action $a$ still takes into account the consumptions making potentially sub-optimal choices. They fix this by picking a slightly different function $\calL_t$.}.
More specifically, they use two online algorithms, one that tries to maximize $\calL_t(a, i)$ over $a$ and one that tries to minimize it over $i$. We follow a similar approach and since the domain of both arguments of $\calL_t(a, i)$ is discrete, we use two \mab algorithms with no-regret guarantees: In round $t$, $\Amax$ chooses an action $A_t$ and $\Amin$ chooses a resource or the number $0$, $I_t$. Then, $\Amax$ receives reward $\calL_t(A_t, I_t)$ and $\Amin$ incurs cost $\calL_t(A_t, I_t)$. We note that the choices of $A_t$ and $I_t$ are made without knowledge of the rewards and consumptions of round $t$. We also note that the player can provide $\Amax$ with bandit information only, i.e., it knows only $\calL(A_t, I_t)$. In contrast, the player can give $\Amin$ full information since it knows the value of $\calL(A_t, i)$ for all $i$. Our full algorithm can be found in Algorithm~\ref{algo:1}.

\begin{algorithm2e}[t]
    \caption{\bwk Algorithm based on Lagrangian maximization/minimization.}
    \label{algo:1}
    \DontPrintSemicolon
    \LinesNumbered
    \KwIn{Maximization algorithm $\Amax$ and minimization algorithm $\Amin$}
    \For{rounds $t=1,2,\ldots$ and while budget lasts}{
        Receive $A_t$ from $\Amax$ and $I_t$ from $\Amin$.\;
        Play action $A_t$.\;
        Receive reward $R_t(A_t)$ and consumptions $C_{t,1}(A_t), \ldots, C_{t,d}(A_t)$.\;
        Update $\Amax$ with reward $\calL_t(A_t, I_t) = R_t(A_t) + \frac{1}{\rho}\One{I_t \ne 0}(\rho - C_{t, I_t}(A_t))$.\;
        Update $\Amin$ with costs $\calL_t(A_t, i) = R_t(A_t) + \frac{1}{\rho}\One{i \ne 0}(\rho - C_{t,i}(A_t))$ for $i\in[d]\cup\{0\}$.\;
    }
\end{algorithm2e}

We will use algorithms $\Amax$ and $\Amin$ that guarantee no-regret with high probability. We use EXP3.P from \cite{DBLP:journals/siamcomp/AuerCFS02} as $\Amax$ which guarantees that for all $\d > 0$ with probability at least $1-\d$ it holds that for all $T'$
\begin{align}\label{eq:prel:Rmax}
    \max_{a\in [K]}\sum_{t=1}^{T'} \calL_t(a, I_t)
    -
    \sum_{t=1}^{T'} \calL_t(A_t, I_t)
    \le
    \Rmax(T, \d)
    :=
    O\left(
        \frac{1}{\rho}
        \sqrt{K T \log(T/\d)}
    \right)
    .
\end{align}

Using Hedge from \cite{DBLP:journals/jcss/FreundS97} as $\Amin$ guarantees that for all $\d > 0$ with probability at least $1-\d$ it holds that for all $T'$
\begin{align}\label{eq:prel:Rmin}
    \sum_{t=1}^{T'} \calL_t(A_t, I_t)
    -
    \min_{i \in [d]\cup\{0\}}\sum_{t=1}^{T'} \calL_t(A_t, i)
    \le
    \Rmin(T, \d)
    :=
    O\left(
        \frac{1}{\rho}
        \sqrt{T \log(Td/\d)}
    \right)
    .
\end{align}

For \bwk with full information, we can use Hedge for $\Amax$ to get the improved regret guarantee $\Rmax(T, \d) = O\left( \frac{1}{\rho} \sqrt{T \log(TK/\d)} \right)$ in \eqref{eq:prel:Rmax}.

%% file: body/6.simple_guarantees.tex
\section{Guarantees of Algorithm \ref{algo:1} in Approximately Stationary \bwk}
\label{sec:guar}

In this section, we prove our guarantee for Algorithm~\ref{algo:1} against a stationary adversary. For any values of $\sR$ and $\sC$ our algorithm achieves the competitive ratio of $1/\rho$ that is guaranteed in Adversarial \bwk. As $\sR$ and $\sC$ increase, the competitive ratio of Algorithm~\ref{algo:1} smoothly improves and becomes $1$ when $\sR = \sC = 1$. In the most interesting range of parameters, when $\rho$ is much smaller than $\sR \sC$ (which also implies that the gap in guarantees between Stochastic and Adversarial \bwk is largest), our algorithm achieves a close to tight $\sR \sC$ fraction of the optimal solution. This is a huge improvement over the $\rho$ fraction that is guaranteed in Adversarial \bwk.

We start with a lemma comparing the rewards achieved by Algorithm~\ref{algo:1} against a distribution of actions whose maximum expected consumption of any resource is at most $\rho$. This lemma is true for any $\sR, \sC$, and adaptive adversary. It also easily proves the two guarantees of \cite{DBLP:conf/icml/CastiglioniCK22} about Stochastic and Adversarial \bwk and extends the second for adaptive adversaries.

\begin{lemma}\label{lem:algo:bound}
    Let $A \in \Delta([K])$ be a distribution of actions such that $\max_{i,t} \Ex[a\sim A]{c_{i,t}(a)} \le \rho$.
    Then for any adversary and $\d > 0$, with probability at least $1-\d$ Algorithm~\ref{algo:1} achieves
    \begin{equation}\label{eq:algo:bound}
        \rew
        \ge
        \sum_{t=1}^T \Ex[a\sim A]{r_t(a)} - \Rmax(T, \d) - \Rmin(T, \d)
    \end{equation}
\end{lemma}

Using Lemma~\ref{lem:algo:bound} it is easy to prove the two guarantees of \cite{DBLP:conf/icml/CastiglioniCK22}. First, if the adversary is stochastic we can prove a $1$ competitive ratio and sublinear regret by noticing that the optimal action $A^*$ in \eqref{eq:optfd_stoch} satisfies the conditions of the lemma. Second, against an adaptive adversary, we notice that the action distribution that plays with probability $\rho$ the best-fixed \textit{unbudgeted} action, and the null action otherwise satisfies Lemma~\ref{lem:algo:bound}. This proves a $1/\rho$ competitive ratio against $\max_a\sum_t r_t(a)$ with high probability, which also proves the same guarantee against $\optfd$ as without a budget, the optimal action is fixed.

\begin{proof}{\bf sketch}
    The lemma's proof is based on the guarantees of the two algorithms, $\Amax$ and $\Amin$, found in \eqref{eq:prel:Rmax} and \eqref{eq:prel:Rmin}, respectively, up to the stopping round of the algorithm, $\TA$.
    On the one hand, we compare $\sum_{t\le\TA} \calL_t(A_t, I_t)$ with $\min_i\sum_{t\le\TA} \calL_t(A_t, i)$: this lower bounds the reward of the algorithm, $\rew$, using an additive term that boosts $\rew$ as $T-\TA$ becomes bigger: if the algorithm runs out of budget fast, the $C_{t, i}(A_t)$ terms in $\calL(A_t, i)$ become larger making this bound better.
    On the other hand, we compare $\sum_{t\le\TA} \calL_t(A_t, I_t)$ with $\sum_{t\le\TA} \Ex[a\sim A]{\calL(a, I_t)}$ (where $A$ is the distribution defined in Lemma~\ref{lem:algo:bound}): this contains the total reward of distribution $A$ up to round $\TA$ and an additive error that is not too high because on expectation $C_{t, i}(a) = c_{t, i}(a)$ and the second term is low by the properties of $A$.
    We defer the detailed proof to Appendix \ref{sec:app:guar}.
\end{proof}

We now move to the main theorem of this section. Our theorem guarantees a fraction of the optimal solution with high probability against an adaptive $(\sR, \sC)$-stationary adversary.

\begin{theorem}\label{thm:guar:guar}
    Against an adaptive $(\sR, \sC)$-stationary adversary, the reward of Algorithm~\ref{algo:1} satisfies for any $\d > 0$ with probability at least $1 - \d$
    \begin{equation*}
        \rew
        \ge
        \left(\rho + \sR (\sC - \rho)^+\right) \optfd
        - \Rmax(T, \d) - \Rmin(T, \d)
        .
    \end{equation*}
\end{theorem}

The proof of the theorem is deferred to Appendix \ref{sec:app:guar}. The idea is to use Lemma~\ref{lem:algo:bound}. To take advantage of \eqref{eq:algo:bound} we need to lower bound the reward of the optimal distribution $A^*$ after its stopping time $T^*$ and upper bound its maximum consumption. These two quantities depend on $T^*$: as $T^*$ becomes larger, both the reward of $A^*$ after $T^*$ and the maximum consumption become smaller (the first depending on $\sR$ and the second on $\sC$). Carefully examining these effects and choosing the $T^*$ (as a function of $\rho,\sR,\sC$) that yields the worst guarantee for the algorithm gets us the theorem.

\begin{remark}
    When $\sR$ and $\sC$ are much larger than $\rho$ a naive approach would be to use an algorithm assuming a fully stochastic setting. Such an algorithm would lead to much weaker results. An algorithm that is based on the classic Arm Elimination algorithm would guarantee only a $\sR^2 \sC^2$ fraction of $\optfd$: it might eliminate an action because it identified it as sub-optimal if it is a bit worse than the one that it has identified as optimal. However, after that round, an $(\sR, \sC)$-stationary adversary might make the previously optimal action worse by a factor of $\sR\sC$ and the sub-optimal one better by the same factor. This would result in a sub-optimality factor of $\sR^2 \sC^2$.
\end{remark}

%% file: body/7.impossibility.tex
\section{Impossibility results for Approximately Stationary \bwk}
\label{sec:imposs}

In this section, we show a bound on the guarantee any algorithm can achieve against an $(\sR,\sC)$-stationary adversary. The bound applies to algorithms that are oblivious to the values of $\sR$ and $\sC$ and guarantee a fraction of at least $\rho$ against an oblivious adversary. Additionally, we get this bound when there is only one resource.
Our theorem has many interesting corollaries.
First, it proves that a $\rho$ fraction of $\optfd$ is the best possible in fully adversarial cases, improving over the result of \cite{DBLP:conf/sigecom/BalseiroG17} that a $\rho$ fraction of the best sequence of actions is the best possible.
Second, when $\rho$ is much smaller than $\sR \sC^2$, it makes the guarantee of Theorem~\ref{thm:guar:guar} approximately tight (in fact if $\rho \le \sR \sC^2$ the two bounds are within a factor of $2$ of each other).
Third, when $\sR = O(\rho)$, it proves that any algorithm can guarantee at most $O(\rho)$ fraction of $\optfd$, making Algorithm~\ref{algo:1} optimal up to constant factors.

\begin{theorem}\label{thm:imposs}
    Fix any algorithm that achieves sublinear regret and an $\a_\rho(\sR, \sC)$ fraction of the optimal solution against an oblivious $(\sR, \sC)$-adversary. If the algorithm is oblivious to the values of $\sR$ and $\sC$, $\rho = \Theta(1)$, and $\a_\rho(0, 0) \ge \rho$ (i.e., guarantees at least a $\rho$ fraction of the optimal solution), then
    \begin{equation*}
        \a_\rho(\sR, \sC)
        \le
        \begin{cases}
            \sR + \rho(1-\sR) ,
             & \textrm{ if } \sR \le \rho
            \\
            2 \sqrt{\sR \rho} - \sR \rho ,
             & \textrm{ if } \rho \le \sR \le \frac{\rho}{\sC^2}
            \\
            \sR \sC + \rho(1/\sC - \sR),
             & \textrm{ if }\sR \ge \frac{\rho}{\sC^2}
        \end{cases}
    \end{equation*}
\end{theorem}

In the example we use to prove the theorem, there are $K+1$ different outcomes (the oblivious adversary selects which outcome the algorithm faces). Additionally, the $T$ rounds are divided into $K$ batches. In outcome $q \in [K]$, the optimal action has positive reward only in the rounds of the $q$-th batch. The algorithm cannot distinguish if it is facing the $q$-th outcome on the $(q-1)$-th batch, so it needs to act conservatively and not spend its resources to guarantee a $\rho$ fraction of the optimal solution. Because of this behavior, any algorithm is guaranteed to miss out on a large part of the optimal solution in the $(K+1)$-instance, which is the one that is $(\sR, \sC)$-stationary.
The proof of the theorem is deferred to Appendix~\ref{sec:app:imposs}.

%% file: body/8.complex_guarantees.tex
\section{Improved guarantee for one resource}
\label{sec:complex}

In this section, we provide an algorithm that improves the guarantee of Algorithm~\ref{algo:1} when $\sC$ is small, that is, the variability in expected consumption is high. The improved guarantee requires an additional assumption, that the sum of the differences in the expected consumptions of the actions is sublinear. With this additional guarantee we can ensure sublinear regret.
The bound in the previous section shows that the guarantee of Theorem~\ref{thm:guar:guar} is close to optimal when $\rho$ and $\sR$ are small and $\sC$ is significantly larger. In contrast, when $\sC \le \rho \le \sR$ Theorem~\ref{thm:guar:guar} provides the adversarial guarantee of $\rho$ while Theorem~\ref{thm:imposs} suggests that much better guarantees may be achievable.

Our improved algorithm uses Algorithm~\ref{algo:1} as a subroutine and a parameter $\Tres$. It runs and restarts Algorithm~\ref{algo:1} every $\Tres$ rounds. We allocate each run of Algorithm~\ref{algo:1} a budget of $\rho \Tres - 1$.
This way the per-round budget in every run is approximately $\rho$.
It also guarantees that every run of the algorithm uses at most $\rho \Tres$ of every resource since Algorithm~\ref{algo:1} would have terminated when going above budget and not get the last item; however, when simulating the algorithm and using the actions it suggests, the player has to know to terminate it herself before it uses more than the desired budget which we achieve by allocating the algorithm $1$ less budget than that.

\begin{algorithm2e}
    \caption{Restarting \bwk Algorithm based on Algorithm \ref{algo:1}}
    \label{algo:2}
    \DontPrintSemicolon
    \LinesNumbered
    \KwIn{Inputs needed for Algorithm~\ref{algo:1} and parameter $\Tres$}

    Split rounds into $\lceil \frac{T}{\Tres} \rceil$ disjoint batches $[T] = \mathcal T_1 \cup \ldots \cup \mathcal T_{\lceil \nicefrac{T}{\Tres} \rceil}$, each batch having $\Tres$ rounds (except maybe for the last one).\;
    \For{each batch $j = 1, \ldots, \lceil \frac{T}{\Tres} \rceil$}{
        Independently of previous rounds, run Algorithm~\ref{algo:1} on rounds $\mathcal T_j$ with budget $\rho |\mathcal T_j| - 1$.\;
    }
\end{algorithm2e}

We now show a lemma for Algorithm~\ref{algo:2}, which will lead to the promised improved guarantee. The main ingredient of our guarantee for Algorithm~\ref{algo:1} was Lemma~\ref{lem:algo:bound} that bounds the reward against a distribution $A$ that satisfies $\max_{i,t} \Ex[a \sim A]{c_{i,t}(a)}\le \rho$. To ensure this condition for an arbitrary distribution $A$, we need to scale it down by playing it only with probability $\rho$ divided by the above maximum over the whole time horizon, and playing the null action with the remaining probability.
The new lemma is structured similarly to Lemma~\ref{lem:algo:bound}: it shows that the reward of Algorithm~\ref{algo:2} is at least the reward of any action distribution across all rounds but scaled down as a function of the consumptions of that distribution. 
In contrast to Lemma~\ref{lem:algo:bound} however, the deterioration of the reward is much more fine-grained. Instead of scaling down the whole reward by the maximum consumption, the reward of each round is scaled down by the \textit{consumptions of that round}.

\begin{lemma}\label{lem:compex}
    For any $A \in \Delta([k])$, Algorithm~\ref{algo:2} guarantees reward that for every $\d > 0$, 
    \begin{equation*}
        \Pr{
        \rew
        \ge
        \sum_{t=1}^T \Ex[a\sim A]{r_t(a)} \min\left\{1,\frac{\rho}{\max_i \Ex[a\sim A]{c_{t,i}(a)}}\right\}
        -
        \reg
        } \ge 1 - \d
    \end{equation*}
    where using $\mathcal E \ge \sum_{t=1}^{T-1} \max_{i\in [d]}\big| \Ex[a\sim A]{c_{t,i}(a) - c_{t+1,i}(a)} \big|$ we have
    \begin{equation*}\label{eq:imp:delta}
        \reg
        =
        \frac{T}{\Tres}\big( \Rmax(\Tres, \d T/\Tres) + \Rmin(\Tres, \d T/\Tres) \big)
        - \frac{\Tres}{\rho}\mathcal E
    \end{equation*}
    If $\Tres = \Theta\left( (\nicefrac{\rho T}{\mathcal{E}})^{\frac 2 3} \right)$ and ignoring dependence on $K, d$ then $\reg = O(T^{\frac 2 3} \mathcal{E}^{\frac 1 3} \log(T/\d) \rho^{-\frac 1 3})$.
\end{lemma}

The proof of the lemma is based on Lemma~\ref{lem:algo:bound}. The reward of Algorithm~\ref{algo:2} in each batch $j$ is at least the reward of distribution $A$ in that batch scaled down by $\max_{i,\tau \in \mathcal T_j} \Ex[a\sim A]{c_{\tau, i}(a)}$. This factor can be improved to $\Ex[a\sim A]{c_{t, i}(a)}$ for any $t\in \mathcal T_j$ by introducing an additive error that depends on the variance of consumptions on that round. Using the condition on $\mathcal E$, this additive error over all batches is sublinear, which proves the lemma. The full proof can be found in Appendix~\ref{sec:app:complex}.

We now present the main result of this section. Using Lemma~\ref{lem:compex} we can get a strictly better bound. Our result is parametric and depends on a parameter $x \in [\rho, 1]$. 

\begin{theorem}\label{thm:complex:guar}
    Against an adaptive $(\sR, \sC)$-stationary adversary, for any $\d > 0$ with probability at least $1 - \d$, Algorithm~\ref{algo:2} guarantees at least a $\a_\rho(\sR, \sC)$ fraction of $\optfd$, where
    \begin{align*}
        \a_\rho(\sR, \sC)
        =
        \min_{x\in [\rho,1]}\left(
        \max\left\{
            \rho,
            x \sC,
            \sR \frac{x}{d+x}
        \right\}
        +
        \max\left\{
            \rho \sR \frac{1-x}{x},
            \sR \sC (1-x)
        \right\}
        \right)
    \end{align*}
    and regret $\reg$ which is sublinear if $\mathcal E/\rho$ is sublinear, where $\reg$ and $\mathcal E$ are defined in Lemma~\ref{lem:compex}.
\end{theorem}

\begin{figure}[t]
    \centering%
    \includegraphics[width=.49\textwidth]{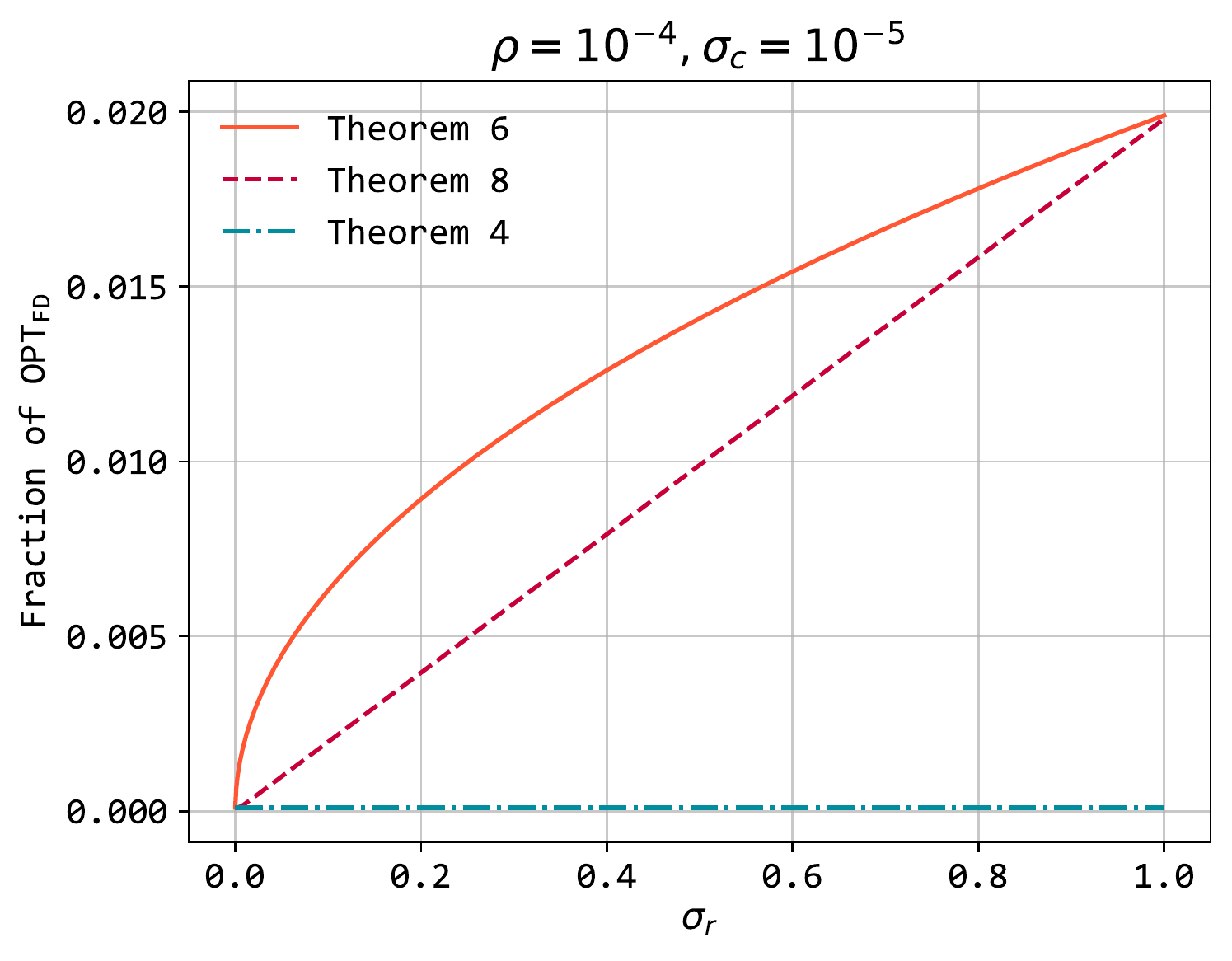}%
    \hfill%
    \includegraphics[width=.49\textwidth]{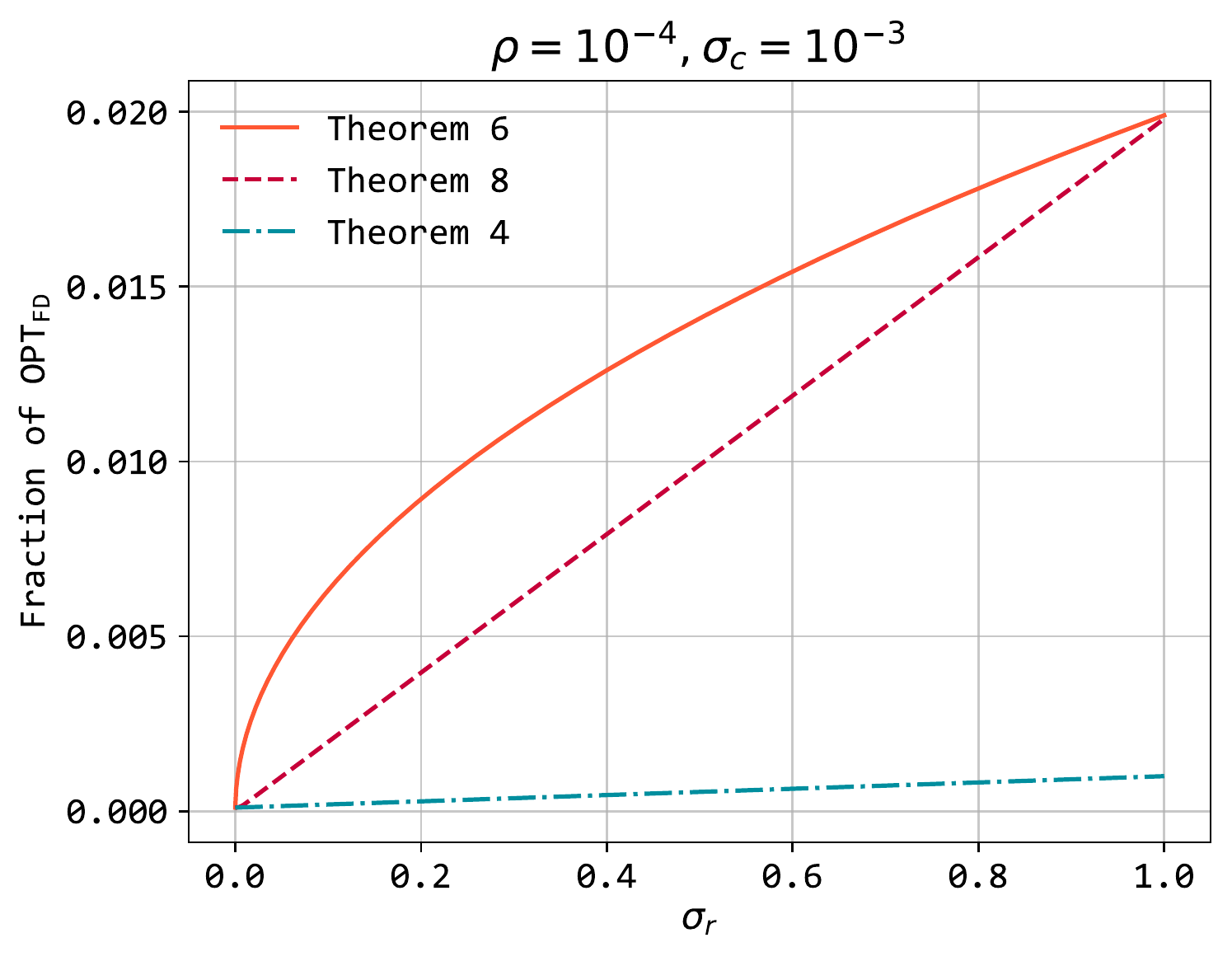}%
    \caption{The bounds of Theorems \ref{thm:guar:guar}, \ref{thm:imposs}, and \ref{thm:complex:guar} on the fraction of the optimal solution for $d = 1$ and small $\rho$ and $\sC$ (mentioned above each graph).}
    \label{fig:numerical}
\end{figure}

\begin{remark}
    The bound of Theorem~\ref{thm:complex:guar} improves the bound of Theorem~\ref{thm:guar:guar} significantly when there is only one resource $d = 1$ and $\sR$ is much larger than $\rho$ and $\sC$.
    For example, if $\sC \le \rho \ll \sR$ then the bound of Theorem~\ref{thm:complex:guar} becomes
    \begin{equation*}
        \a_\rho(\sR, \sC) =
        \begin{cases}
            2\sR(\sqrt \rho - \rho), & \textrm{ if } \sR^2 \ge \rho \\
            \sR^2 + \rho - 2\rho\sR, & \textrm{ if } \sR^2 \le \rho
        \end{cases}
    \end{equation*}
    in which case the bound of Theorem \ref{thm:guar:guar} is $\rho$.
    We also showcase this improvement in Figure~\ref{fig:numerical} using some numerical examples, where we compare the bounds of Theorems~\ref{thm:guar:guar}, \ref{thm:imposs}, and \ref{thm:complex:guar} for small values of $\rho$ and $\sC$ and arbitrary values of $\sR$.

\end{remark}

%% file: appendix/6.simple.tex
\section{Deferred proofs of Section~\ref{sec:guar}}
\label{sec:app:guar}

In this sections we present the deferred proofs for the guarantee of Theorem~\ref{thm:guar:guar}. We first present the proof of Lemma~\ref{lem:algo:bound}.

\begin{proof}\textbf{of Lemma~\ref{lem:algo:bound}}
    We start the proof by reminding the two no-regret guarantees of Algorithm~\ref{algo:1}. \eqref{eq:prel:Rmax} and \eqref{eq:prel:Rmin}, for $T' = \TA$, the algorithm's stopping time imply that for any $\d > 0$ it holds that
    \begin{align}\label{eq:algo:Rmax}
        \Pr{
        \sum_{t=1}^{\TA} \calL_t(A_t, I_t)
        \ge
        \max_{a \in K} \sum_{t=1}^{\TA} \calL_t(a, I_t) -\Rmax(T, \d)
        } \ge 1-\d
        .
    \end{align}
    and
    \begin{align}\label{eq:algo:Rmin}
        \Pr{
        \sum_{t=1}^{\TA} \calL_t(A_t, I_t)
        \le
        \min_{i \in [d]\cup\{0\}} \sum_{t=1}^{\TA} \calL_t(A_t, i) + \Rmin(T, \d)
        } \ge 1-\d
    \end{align}
    
    We start by analyzing the second bound, \eqref{eq:algo:Rmin}. For any $\d_1 > 0$, with probability at least $1 - \d_1$,
    \begin{alignat}{4} \label{eq:5:1}
        \Line{\sum_{t=1}^{\TA} \calL_t(A_t, I_t) - \Rmin(T, \d_1)}
        {\le}{
            \min_{i \in [d]\cup\{0\}} \sum_{t=1}^{\TA} \calL_t(A_t, i)
        }{}
        \nonumber\\
        \Line{}{=}{
            \sum_{t=1}^{\TA} R_t(A_t)
            + \min_{i \in [d]\cup\{0\}} \left\{
                \frac{\One{i > 0}}{\rho}\sum_{t=1}^{\TA}(\rho - C_{t,i}(A_t))
            \right\}
        }{}
        \nonumber\\
        \Line{}{\le}{
            \rew - (T - \TA) + \frac{1}{\rho}
        }{}        
    \end{alignat}
    where the last inequality holds by the following case analysis:
    \begin{itemize}
        \item If $\TA = T$, then, by setting $i = 0$:
        \begin{equation*}
            \min_{i \in [d]\cup\{0\}} \left\{
            \frac{\One{i > 0}}{\rho}\sum_{t=1}^{\TA}(\rho - C_{t,i}(A_t))
            \right\}
            \le
            0
            =
            - (T - \TA)
        \end{equation*}

        \item If $\TA < T$, then the algorithm runs out of some resource $i' \in [d]$ on round $\TA + 1$, meaning that $\sum_{t=1}^{\TA+1} C_{t, i'}(A_t) > \rho T$. By setting $i = i'$ we get
        \begin{equation*}
            \min_{i \in [d]\cup\{0\}} \left\{
            \frac{\One{i > 0}}{\rho}\sum_{t=1}^{\TA}(\rho - C_{t,i}(A_t))
            \right\}
            \le
            \frac{1}{\rho}\sum_{t=1}^{\TA}(\rho - C_{t, i'}(A_t))
            =
            \frac{1}{\rho}\left( \TA \rho - \rho T + 1 \right)
        \end{equation*}
    \end{itemize}

    Now, we analyze \eqref{eq:algo:Rmax}. For any $\d_2 > 0$, with probability at least $1- \d_2$ it holds that 
    \begin{alignat}{4} \label{eq:5:2}
        \Line{
            \sum_{t=1}^{\TA} \calL_t(A_t, I_t) + \Rmax(T, \d_2)
        }{\ge}{
            \max_{a\in [K]} \sum_{t=1}^{\TA} \calL_t(a, I_t)
        }{}
        \nonumber\\
        \Line{}{\ge}{
            \sum_{t=1}^{\TA} \Ex[a\sim A]{\calL_t(a, I_t)}
        }{}
        \nonumber\\
        \Line{}{=}{
            \sum_{t=1}^{\TA} \Ex[a\sim A]{R_t(a)}
            + \sum_{t=1}^{\TA}\frac{\One{I_t > 0}}{\rho} (\rho - \Ex[a\sim A]{C_{t, I_t}(a)})
        }{}
    \end{alignat}

    We now prove a high probability lower bound on the last quantity. Towards that end, we define for every $\tau\in \{0\}\cup[T]$ and $a\in [K]$
    \begin{align*}
        Z_\tau(a)
        =
        \sum_{t=1}^{\tau} \left(R_t(a) - r_t(a)\right)
        + \sum_{t=1}^{\tau}\frac{\One{I_t > 0}}{\rho} \left(c_{t, I_t}(a) - C_{t, I_t}(A)\right)
        .
    \end{align*}

    We notice that for every $a$, $Z_\tau(a)$ is a martingale since
    \begin{align*}
        \Ex{Z_\tau(a) - Z_{\tau - 1}(a) \big| \hist_{\tau-1}}
        =
        \Ex{ R_t(a) - r_t(a)
        + \frac{\One{I_\tau > 0}}{\rho}
          \left( c_{\tau, I_\tau}(a) - C_{\tau, I_\tau}(a) \right) \big| \hist_{\tau-1}}
        =
        0
    \end{align*}
    where the last equality follows from the definitions of $r_t$ and $c_{t, i}$: for all $a$ and all $i$ it holds that $\Ex{ R_t(a) - r_t(a) \big| \hist_{\tau-1}} = 0$ as well as $\Ex{ C_{t, i}(a) - c_{t, i}(a) \big| \hist_{\tau-1}} = 0$.

    Since $Z_\tau(a)$ is a martingale and $|Z_\tau(a) - Z_{\tau-1}(a)| \le 1 + 1/\rho$ we can use the Azuma-Hoeffding inequality and the union bound to get that for any $\d_3 > 0$ with probability at least $1 - \d_3$ it holds for all $T' \in [T]$ and all $a\in[K]$ that
    \begin{align*}
        Z_{T'}(a) - Z_0(a)
        \ge
        - \left( 1 + \frac{1}{\rho} \right) \sqrt{2 T \log(T K /\d_3)}
        \ge
        - \Rmax(T, \d_3)
    \end{align*}

    Combining the above for $T' = \TA$, we can make every $R_t(a)$ and $C_{t,I_t}(a)$ in \eqref{eq:5:2} into $r_t(a)$ and $c_{t,I_t}(a)$, with an additive error of $\Rmax(T, \d_3)$. This yields that with probability at least $1 - \d_2 - \d_3$
    \begin{alignat}{4}\label{eq:5:3}
        \Line{
            \sum_{t=1}^{\TA} \calL_t(A_t, I_t) + \Rmax(T, \d_2)
        }{\ge}{
            \sum_{t=1}^{\TA} \Ex[a\sim A]{r_t(a)}
            + \sum_{t=1}^{\TA}\frac{\One{I_t > 0}}{\rho} (\rho - \Ex[a\sim A]{c_{t, I_t}(a)})
            - \Rmax(T, \d_3)
        }{}
        \nonumber\\
        \Line{}{\ge}{
            \sum_{t=1}^{\TA} \Ex[a\sim A]{r_t(a)}
            - \Rmax(T, \d_3)
        }{}
        \nonumber\\
        \Line{}{\ge}{
            \sum_{t=1}^{T} \Ex[a\sim A]{r_t(a)} - (T - \TA) - \Rmax(T, \d_3)
        }{}
    \end{alignat}
    where in the second inequality we used that $\max_{t,i} \Ex[a\sim A]{c_{t,i}(a)} \le \rho$ and in the last one that $r_t(a) \le 1$.

    Combining \eqref{eq:5:1} and \eqref{eq:5:3}, using the union bound, and setting $\d_1 = \d_2 = \d_3 = \d/3$ we get the desired result with an additive error of $2\Rmax(T, \d/3) + \Rmin(T, \d/3)$. This is the same as the one in the lemma's statement because asymptotically they are the same and by the definition of $\Rmax$ and $\Rmin$ (which are defined using $O(\cdot)$ notation).
\end{proof}

Before proving Theorem~\ref{thm:guar:guar}, we prove two lemmas that show the effects of a $(\sR, \sC)$-stationary adversary to the optimal solution.

The first lemma lower bounds the reward of the optimal distribution of distribution $A^*$ after its stopping time, $T^*$. We need such a bound because we cannot guarantee that the algorithm has gained a large part of $\optfd$ before round $T^*$, especially when $T^*$ is much smaller than $T$. We prove a bound that depends on $\sR$ and $T^*/T$.

\begin{lemma}\label{lem:guar:rewards}
    Let $(A^*, T^*)$ be the optimal solution to \eqref{eq:optfd} and let $x = T^* / T$. Then it holds that
    \begin{align*}
        \sum_{t=T^*+1}^T \Ex[a\sim A^*]{r_t(a)}
        \ge
        \sR \frac{1 - x}{x} \optfd
        .
    \end{align*}
\end{lemma}

Definition~\ref{def:stat_bwk} only states that the rewards of each action are stationary but we  need this for a distribution of actions.

\begin{proof}
    For every $a \in [K]$ let $p(a) = \Pr{A^* = a}$. We have that
    \begin{alignat*}{4}
        \Line{
            \sum_{t=T^* + 1}^T \Ex[a\sim A^*]{r_t(a)}
        }{=}{
            \sum_{a\in [K]} p(a) \sum_{t=T^* + 1}^T r_t(a)
        }{}
        \\
        \Line{}{\ge}{
            \sum_{a\in [K]} p(a) (T - T^*) \min_t r_t(a)
        }{}
        \\
        \Line{}{\ge}{
            \sR \sum_{a\in [K]} p(a) (T - T^*) \max_t r_t(a)
        }{\textrm{$\sR$-stationarity}}
        \\
        \Line{}{\ge}{
            \sR \sum_{a\in [K]} p(a) \frac{T - T^*}{T^*} \sum_{t=1}^{T^*} r_t(a)
        }{}
        \\
        \Line{}{=}{
            \sR \frac{1 - x}{x} \optfd
        }{T^* = x T}
    \end{alignat*}
    which proves the lemma.
\end{proof}

We proceed to prove the second lemma, for the consumption of the optimal action. In contrast to the previous lemma, we prove an upper bound on the maximum consumption across all resources and rounds.

\begin{lemma}\label{lem:guar:costs}
    Let $(A^*, T^*)$ be the optimal solution to \eqref{eq:optfd} and let $x = T^* / T$. Then, for all $i \in [d]$,
    \begin{align*}
        \max_{t} \Ex[a\sim A^*]{c_{t,i}(a)}
        \le
        \frac{\rho}{x \sC}
        .
    \end{align*}
\end{lemma}

To prove this lemma we use the fact that by round $T^*$ the total consumption of any resource under action $A^*$ is at most $T \rho$. We then bound the maximum consumption using $\sC$-stationarity.

\begin{proof}
    For every $a \in [K]$ let $p(a) = \Pr{A^* = a}$. Fix a resource $i\in [d]$. We have that
    \begin{alignat*}{4}
        \Line{
            \max_t \Ex[a\sim A^*]{c_{t,i}(a)}
        }{=}{
            \max_t \sum_{a\in[K]} p(a) c_{t,i}(a)
        }{}
        \\
        \Line{}{\le}{
            \sum_{a\in[K]} p(a) \max_t c_{t,i}(a)    
        }{\textrm{Jensen's inequality}}
        \\
        \Line{}{\le}{
            \frac{1}{\sC} \sum_{a\in[K]} p(a) \min_t c_{t,i}(a)
        }{\textrm{$\sC$-stationary}}
        \\
        \Line{}{\le}{
            \frac{1}{\sC} \sum_{a\in[K]} p(a) \frac{1}{T^*} \sum_{t=1}^{T^*} c_{t,i}(a)
        }{\textrm{minimum less than average}}
        \\
        \Line{}{=}{
            \frac{1}{\sC} \frac{1}{T^*} \sum_{t=1}^{T^*} \Ex[a\sim A^*]{c_{t,i}(a)}
        }{}
        \\
        \Line{}{\le}{
            \frac{1}{\sC} \frac{1}{T^*} T \rho
            =
            \frac{\rho}{x \sC}
        }{\textrm{using \eqref{eq:optfd}}}
    \end{alignat*}

    This proves the lemma.
\end{proof}

We know prove Theorem~\ref{thm:guar:guar}. The proof combines Lemmas~\ref{lem:algo:bound}, \ref{lem:guar:rewards}, and \ref{lem:guar:costs} and guarantees a bound based on $x$. By taking the worst case of over $x$ we get the theorem.

\begin{proof}\textbf{of Theorem~\ref{thm:guar:guar}}
    The theorem trivially holds if $\sC \le \rho$, by the competitive ratio guarantee of $1/\rho$ of the algorithm in purely adversarial settings.

    We are going to use Lemma~\ref{lem:algo:bound} with a carefully chosen distribution of actions. Let $(A^*, T^*)$ be the optimal solution to \eqref{eq:optfd}. Let $x = T^* / T$.
    We set $A$ to be the arm distribution that plays the optimal distribution $A^*$ with probability $\max\{\rho, \sC x \}$ and otherwise the null arm. We notice that for every resource $i$ and round $t$,
    \begin{align*}
        \Ex[a\sim A]{c_{t,i}(a)}
        =
        \max\{\rho, \sC x \} \Ex[a\sim A^*]{c_{t,i}(a)}
        \le
        \max\{\rho, \sC x \} \min\left\{ 1, \frac{\rho}{\sC x }\right\}
        =
        \rho
    \end{align*}
    where in the inequality we used Lemma~\ref{lem:guar:costs} and $c_{t,i}(a) \le 1$. The above allows us to use Lemma~\ref{lem:algo:bound}: for any $\d > 0$ with probability at least $1 - \d$ we have
    \begin{align*}
        \rew + \Rmax(T, \d) + \Rmin(T, \d)
        \ge
        \sum_{t=1}^T \Ex[a\sim A]{r_t(a)}
        =
        \max\left\{\rho, \sC x \right\} \sum_{t=1}^T \Ex[a\sim A^*]{r_t(a)}
    \end{align*}

    Combining the above with Lemma~\ref{lem:guar:rewards} we get that, up to the additive term $\Rmax(T, \d) + \Rmin(T, \d)$, the algorithm guarantees with high probability the following fraction of $\optfd$:
    \begin{align*}
        \max\left\{\rho, \sC x \right\} \left( 1 + \sR \frac{1 - x}{x} \right)
        =
        \max\left\{\rho + \rho \sR \frac{1 - x}{x}, \sR \sC + \sC x (1 - \sR) \right\}
        .
    \end{align*}

    The above term is decreasing in $x$ while $x \le \frac{\rho}{\sC}$ and increasing afterwards. This means that the minimizing $x$ is $\frac{\rho}{\sC}$, which proves the desired competitive ratio.
\end{proof}

%% file: appendix/7.impossibility.tex
\section{Deferred proofs of Section~\ref{sec:imposs}}
\label{sec:app:imposs}

In this section, we present the deferred proof of Theorem~\ref{thm:imposs}.

\begin{proof}\textbf{of Theorem~\ref{thm:imposs}}
    There is only one resource, $d=1$. Let $y$ be a real number such that $\rho \le y < 1$. We create an instance with the null action and another $K$ actions, where $K = 1 + \lceil \frac{1-y}{\rho}\rceil$. Additionally, let $z = \frac{1-y}{K-1}$. We are going to pick $y$ that does not depend on $T$, which entails that $K$ and $z$ also do not depend on $T$. We split the rounds into $K$ batches of rounds. The first batch has $T y$ rounds and the rest have $\frac{T(1-y)}{K-1} = T z \le T \rho$ rounds each. We examine $K + 1$ different outcomes, indexed by $q \in [K+1]$. The adversary decides the value of $q$. We are going to introduce outcome $q = K+1$ later in the proof and initially focus on the first $K$ outcomes.
    
    We now present the reward and consumption of each arm in each batch and outcome, which can also be found in Table~\ref{tab:example}. During the rounds of batch $i\in[K]$, only the $i$-th arm can have positive reward and consumption. Let $\e < 1$ and let $r_t^{(q)}(a), c_t^{(q)}(a)$ denote the reward and consumption of arm $a$ in outcome $q$. The reward and consumption of arm $i = 1$ are
    \begin{equation}\label{eq:imposs:10}
        \left(r_t^{(q)}(1), c_t^{(q)}(1)\right) =
        \begin{cases}
            \left( \e^{K-1} , \frac{\rho}{y} \right) ,
                      & \textrm{ if } t \in 1\textrm{st batch and for any outcome } q
            \\
            (0 , 0) , & \textrm{ otherwise}
        \end{cases}
    \end{equation}
    and the rewards and costs for every arm $2 \ge i \ge K$
    \begin{equation*}
        \left(r_t^{(q)}(i), c_t^{(q)}(i)\right) =
        \begin{cases}
            \left( \e^{K-i} , 1 \right) ,
                      & \textrm{ if } t \in \textrm{$i$-th batch and outcome is } i \le q \le K
            \\
            (0 , 0) , & \textrm{ otherwise}
        \end{cases}
    \end{equation*}

    \begin{table}[h]
    \centering
    \begin{tabular}{lcccc}
        \toprule
        Outcome $q$ & Action $i=1$, batch $i=1$             & $i=2$           & $i=3$           & $3<i\le K$                        \\ \midrule
        $q=1$       & $\left(\e^{K-1}, \frac{\rho}{y}\right)$ & $(0, 0)$        & $(0, 0)$        & $(0, 0)$                     \\ 
        $q=2$       & $\left(\e^{K-1}, \frac{\rho}{y}\right)$ & $(\e^{K-2}, 1)$ & $(0, 0)$        & $(0, 0)$                     \\ 
        $q=3$       & $\left(\e^{K-1}, \frac{\rho}{y}\right)$ & $(\e^{K-2}, 1)$ & $(\e^{K-3}, 1)$ & $(0, 0)$                     \\ 
        $3<q\le K$  & $\left(\e^{K-1}, \frac{\rho}{y}\right)$ & $(\e^{K-2}, 1)$ & $(\e^{K-3}, 1)$ & $(\e^{K-i}, 1)$ if $q\ge i$ else $(0, 0)$
        \\\bottomrule
    \end{tabular}
    \caption{Rewards and consumptions in the first $K$ outcomes. Each cell shows the tuple of reward and consumption of an action $i$ during batch $i$ in the $q$-th outcome. The reward and consumption of action $a \ne i$ in batch $i$ are both always $0$.}
    \label{tab:example}
    \end{table}


    The optimal solution for outcome $2 \le q \le K$ is at least the reward of playing the $q$-th arm with probability $1$, resulting in its reward being
    \begin{equation*}
        \opt_q
        \ge
        \e^{K-q} T z
    \end{equation*}

    Now fix any algorithm which plays the $i$-th arm on batch $i$, $n_i T$ times. Because of the budget constraints, it must hold that
    \begin{equation}\label{eq:imposs:1}
        \frac{\rho}{y} n_1 T + \sum_{i=2}^K n_i T \le T\rho
    \end{equation}
    
    The reward of the algorithm on the $q$-th outcome is
    \begin{equation*}
        \rew_q
        =
        \sum_{j=1}^q \e^{K-j} n_j T
        \le
        \e^{K-q}(n_q + \e) T
    \end{equation*}

    This means that the fraction of the optimal solution the algorithm achieves in the $q$-th outcome for $q \ge 2$ is
    \begin{equation*}
        \frac{\rew_q}{\opt_q}
        \le
        \frac{n_q}{z} + \frac{\e}{z}
    \end{equation*}

    Since $\e, \rho, K$ are constants w.r.t. $T$, we notice that for every $q$ it must hold that $n_q = \Theta(1)$; otherwise, if $n_q = o(1)$, in outcome $q$ the algorithm would get a $0$ fraction of the optimal solution as $T\to\infty$. This means that the above two relations are true when $T \to \infty$, in which case any regret terms disappear. Now we can take $\e\to 0$. Since the algorithm must guarantee a $\rho$ fraction of the optimal solution this yields for $q \ge 2$
    \begin{equation*}
        n_q \ge \rho z
    \end{equation*}
    which makes \eqref{eq:imposs:1}
    \begin{equation}\label{eq:imposs:n1}
        n_1
        \le
        y (1 - z(K-1))
        =
        y^2
    \end{equation}
    where in the last inequality we used the definition of $z = \frac{1-y}{K-1}$.

    Now we introduce the $(K+1)$-th outcome, which is $(\sR,\sC)$-stationary. In this outcome, during the first batch, the rewards and consumptions of the actions are identical to the other outcomes: only action $1$ has positive reward and cost, as defined in \eqref{eq:imposs:10}. All actions have zero rewards and consumptions on the remaining batches, except for action $1$. More specifically, for $t > T y$, action $1$ has
    \begin{equation*}
        \left(r_t^{(K+1)}(1) , c_t^{(K+1)}(1)\right)
        =
        \left(
        \sR\epsilon^{K-1} , \min\left\{1, \frac{\rho}{y}\frac{1}{\sC}\right\}
        \right)
    \end{equation*}
    which we notice is $(\sR,\sC)$-stationary. Let $1 / a = \min\left\{1, \frac{\rho}{y}\frac{1}{\sC}\right\}$.

    In this new outcome, the optimal solution is the same as the $1$st outcome, i.e., play the $1$st arm with probability $1$:
    \begin{equation*}
        \opt_{K+1}        
        =
        \opt_1
        =
        \e^{K-1} T y
    \end{equation*}

    Because all the outcomes are the same during the first batch, the algorithm cannot distinguish which batch it is in. This means it must play the first arm during the first batch $n_1 T$ times, entailing that the remaining budget after the first batch is
    \begin{equation*}
        T\rho - n_1 T \frac{\rho}{y}
    \end{equation*}

    This entails that in the new outcome the number of times the first arm is played after the first batch is at most
    \begin{align*}
        & \left(T\rho - n_1 T \frac{\rho}{y}\right)
        a
    \end{align*}
    which proves that the reward for this outcome is at most
    \begin{alignat*}{3}
        \Line{
            \rew_{K+1}
        }{\le}{
            \e^{K-1} n_1 T
            +
            \sR \e^{K-1}
            \left(T\rho - n_1 T \frac{\rho}{y}\right)
            a
        }{}
        \\
        \Line{}{=}{
            \e^{K-1} T \left(
            \rho \sR a
            +
            n_1 \left(
            1 - \frac{\rho \sR a}{y}
            \right)
            \right)
        }{}
        \\
        \Line{}{\le}{
            \e^{K-1} T \left(
            \rho \sR a
            +
            y^2\left(
                1 - \frac{\rho \sR a}{y}
            \right)
            \right)
        }{\textrm{using \eqref{eq:imposs:n1} and } \frac{\rho \sR a}{y} \le 1}
        \\
        \Line{}{=}{
            \e^{K-1} T \left(
                \rho \sR a
                +
                y^2
                -
                \rho y \sR a
            \right)
        }{}
    \end{alignat*}
    making the fraction of the optimal solution that the algorithm gets at most
    \begin{alignat*}{3}
        \Line{
            \frac{\rew_{K+1}}{\opt_{K+1}}
        }{\le}{
            \frac{
                \rho \sR a
                +
                y^2
                -
                \rho y \sR a
            }{y}
        }{}
        \\
        \Line{}{=}{
            y + \frac{\rho \sR (1 - y)}{y} \a
        }{}
        \\
        \Line{}{=}{
            y + \frac{\rho \sR (1 - y)}{y} \max\left\{ 1 , \frac{y \sC}{\rho} \right\}
        }{}
    \end{alignat*}

    Similar to before, because the above expression does not depend on $T$, it also holds as $T\to\infty$. This makes any regret terms disappear, making the above the inverse of the competitive ratio.
    Minimizing it over $y$ we set
    \begin{equation*}
        y =
        \begin{cases}
            \rho,
             & \textrm{ if } \sR \le \rho
            \\
            \sqrt{\rho \sR},
             & \textrm{ if } \rho \le \sR \le \frac{\rho}{\sC^2}
            \\
            \frac{\rho}{\sC} ,
             & \textrm{ if } \sR \ge \frac{\rho}{\sC^2}
        \end{cases}
    \end{equation*}
    which proves the desired bound simply by substituting.
\end{proof}

%% file: appendix/8.complex.tex
\section{Deferred proofs of Section~\ref{sec:complex}}
\label{sec:app:complex}

In this section, we present the deferred proofs of Section~\ref{sec:complex}. We start with Lemma~\ref{lem:compex}.

\begin{proof}\textbf{of Lemma~\ref{lem:compex}}
    For ease of notation, for the distribution mentioned in the lemma's statement $A$, we use $r_t(A) = \Ex[a\sim A]{r_t(a)}$ and $c_{t,i}(A) = \Ex[a\sim A]{c_{t,i}(a)}$.
    
    Fix a batch $j$. Let $\mathcal T_j$ be the rounds of that batch and $\rew_j$ be the total reward of the algorithm during it. The per-round budget during batch $j$ is $\frac{\rho \Tres - 1}{\Tres} = \rho - 1/\Tres$. We use Lemma~\ref{lem:algo:bound}, in which we use the distribution of actions that plays $A$ (as defined in the description of Lemma~\ref{lem:compex}) with probability $\min_i\min\left\{1, \frac{\rho}{\max_{t\in T_j} c_{t,i}(A)}\right\} - 1/\Tres$ and the null arm otherwise.
    This proves that with probability $1 - \d$,
    \begin{alignat}{3}\label{eq:complex:0}
        \Line{
            \rew_j
        }{\ge}{
            \left(\min_i\min\left\{1, \frac{\rho}{\max_{t\in \mathcal T_j} c_{t,i}(A)}\right\} - 1 \right)
            \sum_{t\in T_j} r_t(A)
            - \Rmax(\Tres, \d) - \Rmin(\Tres, \d)
        }{}
        \nonumber\\
        \Line{}{\ge}{
            \min_i\min\left\{1, \frac{\rho}{\max_{t\in \mathcal T_j} c_{t,i}(A)}\right\} \sum_{t\in T_j} r_t(A) - 1
            - \Rmax(\Tres, \d) - \Rmin(\Tres, \d)
        }{}
        \nonumber\\
        \Line{}{=}{
            \min_i\min\left\{1, \frac{\rho}{\max_{t\in \mathcal T_j} c_{t,i}(A)}\right\} \sum_{t\in T_j} r_t(A)
            - \Rmax(\Tres, \d) - \Rmin(\Tres, \d)
        }{}
    \end{alignat}
    where in the last equality we used the fact that $\Rmin(\Tres, \d)$ is asymptotically bigger than $1$.

    Let for every $i \in [d]$
    \begin{align*}
        \mathcal{E}_{ji}
        =
        \max_{t\in \mathcal T_j}c_{t,i}(A) - \min_{t\in \mathcal T_j}c_{t,i}(A)
        .
    \end{align*}
    
    We are going to prove that for any $t\in \mathcal T_j$ and $i\in [d]$ it holds that
    \begin{equation}\label{eq:complex:1}
        \min\left\{1, \frac{\rho}{\max_{\tau\in \mathcal T_j} c_{\tau, i}(A)}\right\}
        \ge
        \min\left\{1, \frac{\rho}{c_{t,i}(A)}\right\} - \frac{1}{\rho}\mathcal{E}_{ji}
    \end{equation}
    
    Combining \eqref{eq:complex:0} and \eqref{eq:complex:1} we get that with probability at least $1-\d$
    \begin{equation*}
        \rew_j + \Rmax(\Tres, \d) + \Rmin(\Tres, \d)
        \ge
        \sum_{t\in T_j} r_t(A) \min_i\min\left\{1, \frac{\rho}{c_{t,i}(A)}\right\}
        - \frac{\Tres}{\rho} \max_i \mathcal{E}_{ji}
    \end{equation*}

    With probability $1- \d \Tres / T$ the above inequality holds for all $j$; summing over all $j$ gets us
    \begin{equation*}
        \rew + \frac{T}{\Tres}\big( \Rmax(\Tres, \d) + \Rmin(\Tres, \d) \big)
        \ge
        \sum_{t=1}^T r_t(A) \min_i\min\left\{1, \frac{\rho}{c_{t,i}(A)}\right\}
        -  \frac{\Tres}{\rho}\sum_j\max_i \mathcal{E}_{ji}
    \end{equation*}
    which proves what we want by noticing that $\mathcal{E} \ge \sum_j\max_i \mathcal{E}_{ji}$ and substituting $\d$.

    Now we need to prove \eqref{eq:complex:1}. Fix $i$ and $j$. We have that
    \begin{equation*}
        \min\left\{1, \frac{\rho}{\max_{\tau\in \mathcal T_j} c_{\tau, i}(A)}\right\} + \frac{\mathcal{E}_{ji}}{\rho}
        =
        \min\left\{1, \frac{\rho}{\min_{\tau\in T_j} c_{\tau,i}(A) + \mathcal{E}_{ji}}\right\} + \frac{\mathcal{E}_{ji}}{\rho}
    \end{equation*}

    Let $c = \min_{\tau\in T_j} c_{\tau,i}(A)$. We notice that the above r.h.s. is increasing in $\mathcal{E}_{ji}$; this is obvious for $\rho > c + \mathcal{E}_{ji}$ and if $c + \mathcal{E}_{ji} \ge \rho$ the derivative is $\frac{(c + \mathcal{E}_{ji})^2 - \rho^2}{\rho(c + \mathcal{E}_{ji})} \ge 0$. This means that, because $\mathcal{E}_{ji} \ge 1 - c$, it holds that
    \begin{equation*}
        \min\left\{1, \frac{\rho}{c + \mathcal{E}_{ji}}\right\} + \frac{\mathcal{E}_{ji}}{\rho}
        \ge
        \min\left\{1, \rho\right\} + \frac{1-c}{\rho}
        \ge
        \rho + 1-c
        =
        (1-c) + c \frac{\rho}{c}
        \ge
        \min\left\{1, \frac{\rho}{c}\right\}
    \end{equation*}
    where in the last inequality we used the fact that the min is less than the weighted average. \eqref{eq:complex:1} follows by noticing that for any $t \in \mathcal T_j$, $c \le c_{t,i}(A)$.
\end{proof}

Now we prove Theorem \ref{thm:complex:guar}.

\begin{proof}\textbf{of Theorem \ref{thm:complex:guar}}
    Fix the solution of \eqref{eq:optfd}, $(A^*, T^*)$ and let $x = T^* / T$. We are going to use Lemma~\ref{lem:compex} by setting $A = A^*$. To get a bound on the competitive ratio, we need to lower bound
    \begin{align}\label{eq:complex:11}
        \sum_{t=1}^T \Ex[a\sim A^*]{r_t(a)} \min\left\{1,\frac{\rho}{\max_i \Ex[a\sim A^*]{c_{t,i}(a)}}\right\}
    \end{align}

    First we prove a bound on $\min_t \Ex[a\sim A^*]{r_t(a)}$. The bound we prove is implicitly proven inside the proof of Lemma~\ref{lem:guar:rewards}, but we include it for completeness:
    \begin{alignat}{3}\label{eq:complex:12}
        \Line{
            \min_{t} \Ex[a\sim A^*]{r_t(a)}
        }{=}{
            \min_{t}\sum_{a=1}^K \Pr{A^* = a} r_t(a)
        }{}
        \nonumber\\
        \Line{}{\ge}{
            \sum_{a=1}^k \Pr{A^* = a} \min_t  r_t(a)
        }{}
        \nonumber\\
        \Line{}{\ge}{
            \sR \sum_{a=1}^K \Pr{A^* = a} \max_t r_t(a)
        }{}
        \nonumber\\
        \Line{}{\ge}{
            \frac{\sR}{T^*} \sum_{a=1}^K \Pr{A^* = a} \sum_{t=1}^{T^*} r_t(a)
        }{}
        \nonumber\\
        \Line{}{=}{
            \frac{\sR}{x T} \optfd
        }{}
    \end{alignat}

    Let $c_t = \max_{i} \Ex[a\sim A^*]{c_{t, i}(a)}$ and $\bar c = \max_t c_t$. We focus on the case where $\bar c \ge \rho$, as otherwise \eqref{eq:complex:11} is at least $\optfd$. We first focus on the sum of \eqref{eq:complex:11} for $t > T^*$:
    \begin{align}\label{eq:complex:13}
        \sum_{t=T^*+1}^T \Ex[a\sim A^*]{r_t(a)} \min\left\{1,\frac{\rho}{c_t}\right\}
        \ge
        \frac{\rho}{\bar c} T (1-x)\min_t \Ex[a\sim A^*]{r_t(a)}
        \ge
        \frac{\rho}{\bar c} \sR \frac{1-x}{x} \optfd
    \end{align}
    where in the last inequality we used \eqref{eq:complex:12}.

    To analyze the sum in \eqref{eq:complex:11} for $t \le T^*$ we take two cases. First, we have that
    \begin{align}\label{eq:complex:14}
        \sum_{t=1}^{T^*} \Ex[a\sim A^*]{r_t(a)} \min\left\{1,\frac{\rho}{c_t}\right\}
        \ge
        \frac{\rho}{\bar c}\sum_{t=1}^{T^*} \Ex[a\sim A^*]{r_t(a)}
        =
        \frac{\rho}{\bar c} \optfd
    \end{align}

    Second, we have that
    \begin{alignat}{3}\label{eq:complex:15}
        \Line{
            \sum_{t=1}^{T^*} \Ex[a\sim A^*]{r_t(a)} \min\left\{1,\frac{\rho}{c_t}\right\}
        }{\ge}{
            \left( \min_{t}\Ex[a\sim A^*]{r_t(a)} \right)
            \sum_{t=1}^{T^*} \min\left\{1,\frac{\rho}{c_t}\right\}
        }{}
        \nonumber\\
        \Line{}{\ge}{
            \left( \min_{t}\Ex[a\sim A^*]{r_t(a)} \right)
            \frac{\rho {T^*}^2}{\sum_{t=1}^{T^*} \max\{ \rho, c_t \}}
        }{\textrm{AM-HM inequality}}
        \nonumber\\
        \Line{}{\ge}{
            \left( \min_{t}\Ex[a\sim A^*]{r_t(a)} \right)
            \frac{\rho {T^*}^2}{\sum_{t=1}^{T^*} (c_t + \rho) }
        }{}
        \nonumber\\
        \Line{}{\ge}{
            \left( \min_{t}\Ex[a\sim A^*]{r_t(a)} \right)
            \frac{\rho {T^*}^2}{d \rho T + \rho T^* }
        }{\forall i: \sum_{t=1}^{T^*} c_{t,i}(A^*) \le \rho T}
        \nonumber\\
        \Line{}{=}{
            \min_{t} \Ex[a\sim A^*]{r_t(a)}
            \frac{x^2 T}{d + x }
        }{T^* = x T}
        \nonumber\\
        \Line{}{=}{
            \sR
            \frac{x}{d + x}
            \optfd
        }{ \textrm{using \eqref{eq:complex:12}} }
    \end{alignat}

    Now we combine \eqref{eq:complex:13}, \eqref{eq:complex:14}, and \eqref{eq:complex:15} to get
    \begin{align*}
        \sum_{t=1}^T \Ex[a\sim A^*]{r_t(a)} \min\left\{1,\frac{\rho}{\max_i c_{t,i}(A^*)}\right\}
        \ge
        \max\left\{
            \frac{\rho}{\bar c}
            ,
            \sR \frac{x}{d + x}
        \right\} \optfd
        +
        \frac{\rho}{\bar c} \sR \frac{1-x}{x} \optfd
    \end{align*}
    using Lemma~\ref{lem:guar:costs} and that $\bar c \le 1$ it holds that $\bar c \le \min\{ 1, \nicefrac{\rho}{x\sC} \}$ which we use to get the desired bound.
\end{proof}

%% file: ref.bib
@inproceedings{DBLP:conf/focs/ImmorlicaSSS19,
  author    = {Nicole Immorlica and
               Karthik Abinav Sankararaman and
               Robert E. Schapire and
               Aleksandrs Slivkins},
  editor    = {David Zuckerman},
  title     = {Adversarial Bandits with Knapsacks},
  booktitle = {60th {IEEE} Annual Symposium on Foundations of Computer Science, {FOCS}
               2019, Baltimore, Maryland, USA, November 9-12, 2019},
  pages     = {202--219},
  publisher = {{IEEE} Computer Society},
  year      = {2019},
  url       = {https://doi.org/10.1109/FOCS.2019.00022},
  doi       = {10.1109/FOCS.2019.00022},
  bibsource = {dblp computer science bibliography, https://dblp.org}
}

@inproceedings{DBLP:conf/icml/CastiglioniCK22,
  author    = {Matteo Castiglioni and
               Andrea Celli and
               Christian Kroer},
  editor    = {Kamalika Chaudhuri and
               Stefanie Jegelka and
               Le Song and
               Csaba Szepesv{\'{a}}ri and
               Gang Niu and
               Sivan Sabato},
  title     = {Online Learning with Knapsacks: the Best of Both Worlds},
  booktitle = {International Conference on Machine Learning, {ICML} 2022, 17-23 July
               2022, Baltimore, Maryland, {USA}},
  series    = {Proceedings of Machine Learning Research},
  volume    = {162},
  pages     = {2767--2783},
  publisher = {{PMLR}},
  year      = {2022},
  url       = {https://proceedings.mlr.press/v162/castiglioni22a.html},
  bibsource = {dblp computer science bibliography, https://dblp.org}
}

@article{DBLP:journals/jcss/FreundS97,
  author    = {Yoav Freund and
               Robert E. Schapire},
  title     = {A Decision-Theoretic Generalization of On-Line Learning and an Application
               to Boosting},
  journal   = {J. Comput. Syst. Sci.},
  volume    = {55},
  number    = {1},
  pages     = {119--139},
  year      = {1997},
  url       = {https://doi.org/10.1006/jcss.1997.1504},
  doi       = {10.1006/jcss.1997.1504},
  bibsource = {dblp computer science bibliography, https://dblp.org}
}

@article{DBLP:journals/siamcomp/AuerCFS02,
  author    = {Peter Auer and
               Nicol{\`{o}} Cesa{-}Bianchi and
               Yoav Freund and
               Robert E. Schapire},
  title     = {The Nonstochastic Multiarmed Bandit Problem},
  journal   = {{SIAM} J. Comput.},
  volume    = {32},
  number    = {1},
  pages     = {48--77},
  year      = {2002},
  url       = {https://doi.org/10.1137/S0097539701398375},
  doi       = {10.1137/S0097539701398375},
  bibsource = {dblp computer science bibliography, https://dblp.org}
}

@article{DBLP:journals/sigecom/Slivkins20,
  author    = {Aleksandrs Slivkins},
  title     = {Book announcement: Introduction to Multi-Armed Bandits},
  journal   = {SIGecom Exch.},
  volume    = {18},
  number    = {1},
  pages     = {28--30},
  year      = {2020},
  url       = {https://doi.org/10.1145/3440959.3440965},
  doi       = {10.1145/3440959.3440965},
  bibsource = {dblp computer science bibliography, https://dblp.org}
}

@article{DBLP:journals/ior/BesbesGZ15,
  author    = {Omar Besbes and
               Yonatan Gur and
               Assaf Zeevi},
  title     = {Non-Stationary Stochastic Optimization},
  journal   = {Oper. Res.},
  volume    = {63},
  number    = {5},
  pages     = {1227--1244},
  year      = {2015},
  url       = {https://doi.org/10.1287/opre.2015.1408},
  doi       = {10.1287/opre.2015.1408},
  bibsource = {dblp computer science bibliography, https://dblp.org}
}

@inproceedings{DBLP:conf/nips/GurZB14,
  author    = {Yonatan Gur and
               Assaf Zeevi and
               Omar Besbes},
  editor    = {Zoubin Ghahramani and
               Max Welling and
               Corinna Cortes and
               Neil D. Lawrence and
               Kilian Q. Weinberger},
  title     = {Stochastic Multi-Armed-Bandit Problem with Non-stationary Rewards},
  booktitle = {Advances in Neural Information Processing Systems 27: Annual Conference
               on Neural Information Processing Systems 2014, December 8-13 2014,
               Montreal, Quebec, Canada},
  pages     = {199--207},
  year      = {2014},
  url       = {https://proceedings.neurips.cc/paper/2014/hash/903ce9225fca3e988c2af215d4e544d3-Abstract.html},
  bibsource = {dblp computer science bibliography, https://dblp.org}
}

@inproceedings{DBLP:conf/icml/BalseiroLM20,
  author    = {Santiago R. Balseiro and
               Haihao Lu and
               Vahab S. Mirrokni},
  title     = {Dual Mirror Descent for Online Allocation Problems},
  booktitle = {Proceedings of the 37th International Conference on Machine Learning,
               {ICML} 2020, 13-18 July 2020, Virtual Event},
  series    = {Proceedings of Machine Learning Research},
  volume    = {119},
  pages     = {613--628},
  publisher = {{PMLR}},
  year      = {2020},
  url       = {http://proceedings.mlr.press/v119/balseiro20a.html},
  bibsource = {dblp computer science bibliography, https://dblp.org}
}

@inproceedings{DBLP:conf/focs/BadanidiyuruKS13,
  author    = {Ashwinkumar Badanidiyuru and
               Robert Kleinberg and
               Aleksandrs Slivkins},
  title     = {Bandits with Knapsacks},
  booktitle = {54th Annual {IEEE} Symposium on Foundations of Computer Science, {FOCS}
               2013, 26-29 October, 2013, Berkeley, CA, {USA}},
  pages     = {207--216},
  publisher = {{IEEE} Computer Society},
  year      = {2013},
  url       = {https://doi.org/10.1109/FOCS.2013.30},
  doi       = {10.1109/FOCS.2013.30},
  bibsource = {dblp computer science bibliography, https://dblp.org}
}

@inproceedings{DBLP:conf/sigecom/AgrawalD14,
  author    = {Shipra Agrawal and
               Nikhil R. Devanur},
  editor    = {Moshe Babaioff and
               Vincent Conitzer and
               David A. Easley},
  title     = {Bandits with concave rewards and convex knapsacks},
  booktitle = {{ACM} Conference on Economics and Computation, {EC} '14, Stanford
               , CA, USA, June 8-12, 2014},
  pages     = {989--1006},
  publisher = {{ACM}},
  year      = {2014},
  url       = {https://doi.org/10.1145/2600057.2602844},
  doi       = {10.1145/2600057.2602844},
  bibsource = {dblp computer science bibliography, https://dblp.org}
}

@inproceedings{DBLP:conf/aistats/SankararamanS18,
  author    = {Karthik Abinav Sankararaman and
               Aleksandrs Slivkins},
  editor    = {Amos J. Storkey and
               Fernando P{\'{e}}rez{-}Cruz},
  title     = {Combinatorial Semi-Bandits with Knapsacks},
  booktitle = {International Conference on Artificial Intelligence and Statistics,
               {AISTATS} 2018, 9-11 April 2018, Playa Blanca, Lanzarote, Canary Islands,
               Spain},
  series    = {Proceedings of Machine Learning Research},
  volume    = {84},
  pages     = {1760--1770},
  publisher = {{PMLR}},
  year      = {2018},
  url       = {http://proceedings.mlr.press/v84/sankararaman18a.html},
  bibsource = {dblp computer science bibliography, https://dblp.org}
}

@inproceedings{DBLP:conf/colt/BadanidiyuruLS14,
  author    = {Ashwinkumar Badanidiyuru and
               John Langford and
               Aleksandrs Slivkins},
  editor    = {Maria{-}Florina Balcan and
               Vitaly Feldman and
               Csaba Szepesv{\'{a}}ri},
  title     = {Resourceful Contextual Bandits},
  booktitle = {Proceedings of The 27th Conference on Learning Theory, {COLT} 2014,
               Barcelona, Spain, June 13-15, 2014},
  series    = {{JMLR} Workshop and Conference Proceedings},
  volume    = {35},
  pages     = {1109--1134},
  publisher = {JMLR.org},
  year      = {2014},
  url       = {http://proceedings.mlr.press/v35/badanidiyuru14.html},
  bibsource = {dblp computer science bibliography, https://dblp.org}
}

@inproceedings{DBLP:conf/ijcai/RangiFT19,
  author    = {Anshuka Rangi and
               Massimo Franceschetti and
               Long Tran{-}Thanh},
  editor    = {Sarit Kraus},
  title     = {Unifying the Stochastic and the Adversarial Bandits with Knapsack},
  booktitle = {Proceedings of the Twenty-Eighth International Joint Conference on
               Artificial Intelligence, {IJCAI} 2019, Macao, China, August 10-16,
               2019},
  pages     = {3311--3317},
  publisher = {ijcai.org},
  year      = {2019},
  url       = {https://doi.org/10.24963/ijcai.2019/459},
  doi       = {10.24963/ijcai.2019/459},
  bibsource = {dblp computer science bibliography, https://dblp.org}
}

@inproceedings{DBLP:conf/colt/Kesselheim020,
  author    = {Thomas Kesselheim and
               Sahil Singla},
  editor    = {Jacob D. Abernethy and
               Shivani Agarwal},
  title     = {Online Learning with Vector Costs and Bandits with Knapsacks},
  booktitle = {Conference on Learning Theory, {COLT} 2020, 9-12 July 2020, Virtual
               Event [Graz, Austria]},
  series    = {Proceedings of Machine Learning Research},
  volume    = {125},
  pages     = {2286--2305},
  publisher = {{PMLR}},
  year      = {2020},
  url       = {http://proceedings.mlr.press/v125/kesselheim20a.html},
  bibsource = {dblp computer science bibliography, https://dblp.org}
}

@inproceedings{DBLP:conf/sigecom/BalseiroG17,
  author    = {Santiago R. Balseiro and
               Yonatan Gur},
  editor    = {Constantinos Daskalakis and
               Moshe Babaioff and
               Herv{\'{e}} Moulin},
  title     = {Learning in Repeated Auctions with Budgets: Regret Minimization and
               Equilibrium},
  booktitle = {Proceedings of the 2017 {ACM} Conference on Economics and Computation,
               {EC} '17, Cambridge, MA, USA, June 26-30, 2017},
  pages     = {609},
  publisher = {{ACM}},
  year      = {2017},
  url       = {https://doi.org/10.1145/3033274.3084088},
  doi       = {10.1145/3033274.3084088},
  bibsource = {dblp computer science bibliography, https://dblp.org}
}

@article{DBLP:journals/ftopt/Hazan16,
  author    = {Elad Hazan},
  title     = {Introduction to Online Convex Optimization},
  journal   = {Found. Trends Optim.},
  volume    = {2},
  number    = {3-4},
  pages     = {157--325},
  year      = {2016},
  url       = {https://doi.org/10.1561/2400000013},
  doi       = {10.1561/2400000013},
  bibsource = {dblp computer science bibliography, https://dblp.org}
}

@inproceedings{DBLP:journals/jmlr/BubeckS12,
  author    = {S{\'{e}}bastien Bubeck and
               Aleksandrs Slivkins},
  editor    = {Shie Mannor and
               Nathan Srebro and
               Robert C. Williamson},
  title     = {The Best of Both Worlds: Stochastic and Adversarial Bandits},
  booktitle = {{COLT} 2012 - The 25th Annual Conference on Learning Theory, June
               25-27, 2012, Edinburgh, Scotland},
  series    = {{JMLR} Proceedings},
  volume    = {23},
  pages     = {42.1--42.23},
  publisher = {JMLR.org},
  year      = {2012},
  url       = {http://proceedings.mlr.press/v23/bubeck12b/bubeck12b.pdf},
  bibsource = {dblp computer science bibliography, https://dblp.org}
}

@inproceedings{DBLP:conf/stoc/LykourisML18,
  author    = {Thodoris Lykouris and
               Vahab S. Mirrokni and
               Renato Paes Leme},
  editor    = {Ilias Diakonikolas and
               David Kempe and
               Monika Henzinger},
  title     = {Stochastic bandits robust to adversarial corruptions},
  booktitle = {Proceedings of the 50th Annual {ACM} {SIGACT} Symposium on Theory
               of Computing, {STOC} 2018, Los Angeles, CA, USA, June 25-29, 2018},
  pages     = {114--122},
  publisher = {{ACM}},
  year      = {2018},
  url       = {https://doi.org/10.1145/3188745.3188918},
  doi       = {10.1145/3188745.3188918},
  bibsource = {dblp computer science bibliography, https://dblp.org}
}

@article{DBLP:journals/tit/TekinL12,
  author    = {Cem Tekin and
               Mingyan Liu},
  title     = {Online Learning of Rested and Restless Bandits},
  journal   = {{IEEE} Trans. Inf. Theory},
  volume    = {58},
  number    = {8},
  pages     = {5588--5611},
  year      = {2012},
  url       = {https://doi.org/10.1109/TIT.2012.2198613},
  doi       = {10.1109/TIT.2012.2198613},
  bibsource = {dblp computer science bibliography, https://dblp.org}
}

@inproceedings{DBLP:conf/nips/WangHL20,
  author    = {Siwei Wang and
               Longbo Huang and
               John C. S. Lui},
  editor    = {Hugo Larochelle and
               Marc'Aurelio Ranzato and
               Raia Hadsell and
               Maria{-}Florina Balcan and
               Hsuan{-}Tien Lin},
  title     = {Restless-UCB, an Efficient and Low-complexity Algorithm for Online
               Restless Bandits},
  booktitle = {Advances in Neural Information Processing Systems 33: Annual Conference
               on Neural Information Processing Systems 2020, NeurIPS 2020, December
               6-12, 2020, virtual},
  year      = {2020},
  url       = {https://proceedings.neurips.cc/paper/2020/hash/89ae0fe22c47d374bc9350ef99e01685-Abstract.html},
  bibsource = {dblp computer science bibliography, https://dblp.org}
}

@article{DBLP:journals/jacm/DevanurJSW19,
  author       = {Nikhil R. Devanur and
                  Kamal Jain and
                  Balasubramanian Sivan and
                  Christopher A. Wilkens},
  title        = {Near Optimal Online Algorithms and Fast Approximation Algorithms for
                  Resource Allocation Problems},
  journal      = {J. {ACM}},
  volume       = {66},
  number       = {1},
  pages        = {7:1--7:41},
  year         = {2019},
  url          = {https://doi.org/10.1145/3284177},
  doi          = {10.1145/3284177},
  bibsource    = {dblp computer science bibliography, https://dblp.org}
}

@inproceedings{DBLP:conf/nips/LiuJL22,
  author       = {Shang Liu and
                  Jiashuo Jiang and
                  Xiaocheng Li},
  title        = {Non-stationary Bandits with Knapsacks},
  booktitle    = {NeurIPS},
  year         = {2022},
  url          = {http://papers.nips.cc/paper\_files/paper/2022/hash/69469da823348084ca8933368ecbf676-Abstract-Conference.html},
  bibsource    = {dblp computer science bibliography, https://dblp.org}
}

@misc{slivkins2023contextual,
    title={Contextual Bandits with Packing and Covering Constraints: A Modular Lagrangian Approach via Regression}, 
    author={Aleksandrs Slivkins and Karthik Abinav Sankararaman and Dylan J. Foster},
    year={2023},
    eprint={2211.07484},
    archivePrefix={arXiv},
    primaryClass={cs.LG}
}

@inproceedings{DBLP:conf/nips/KumarK22,
  author       = {Raunak Kumar and
                  Robert Kleinberg},
  title        = {Non-monotonic Resource Utilization in the Bandits with Knapsacks Problem},
  booktitle    = {NeurIPS},
  year         = {2022},
  url          = {http://papers.nips.cc/paper\_files/paper/2022/hash/7a62d9a4c03377d1175b8859b4cc16d4-Abstract-Conference.html},
  bibsource    = {dblp computer science bibliography, https://dblp.org}
}
